\documentclass[manuscript,screen]{acmart}

\usepackage{amsmath,multirow}
\usepackage{graphicx}
\usepackage{algorithm}
\usepackage{algorithmic}
\usepackage{subfigure,color}
\newcommand{\hao}[1]{\textcolor{black}{#1}}
\newcommand{\tht}[2]{\begin{tabular}{@{}#1@{}}#2\end{tabular}}
\usepackage{gensymb,graphics,subfigure,inputenc}
\graphicspath{{Figures/}}
\usepackage{textcomp,subfigure,multirow,upgreek}
\usepackage{booktabs}
\usepackage{color,mdwlist}
\usepackage{lineno,hyperref}

\AtBeginDocument{%
  \providecommand\BibTeX{{%
    \normalfont B\kern-0.5em{\scshape i\kern-0.25em b}\kern-0.8em\TeX}}}

\setcopyright{acmcopyright}
\copyrightyear{2018}
\acmYear{2018}
\acmDOI{10.1145/1122445.1122456}

\acmConference[Woodstock '18]{Woodstock '18: ACM Symposium on Neural
  Gaze Detection}{June 03--05, 2018}{Woodstock, NY}
\acmBooktitle{Woodstock '18: ACM Symposium on Neural Gaze Detection,
  June 03--05, 2018, Woodstock, NY}
\acmPrice{15.00}
\acmISBN{978-1-4503-XXXX-X/18/06}

\begin{document}

\title{Asymmetric GANs for Image-to-Image Translation}

\author{Hao Tang}
\email{hao.tang@vision.ee.ethz.ch}
\authornotemark[1]
\affiliation{%
  \institution{ETH Zurich}
%  \streetaddress{P.O. Box 1212}
\country{Switzerland}
}

\author{Nicu Sebe}
\affiliation{%
  \institution{University of Trento}
%  \streetaddress{1 Th{\o}rv{\"a}ld Circle}
  \country{Italy}}
%\email{larst@affiliation.org}

\renewcommand{\shortauthors}{Hao Tang and Nicu Sebe}

\begin{abstract}
	Existing models for unsupervised image translation with Generative Adversarial Networks (GANs) can learn the mapping from the source domain to the target domain using a cycle-consistency loss. However, these methods always adopt a symmetric network architecture to learn both forward and backward cycles. Because of the task complexity and cycle input difference between the source and target domains, the inequality in bidirectional forward-backward cycle translations is significant and the amount of information between two domains is different. In this paper, we analyze the limitation of existing symmetric GANs in asymmetric translation tasks, and propose an AsymmetricGAN model with both translation and reconstruction generators of unequal sizes and different parameter-sharing strategy to adapt to the asymmetric need in both unsupervised and supervised image translation tasks. Moreover,  the training stage of existing methods has the common problem of model collapse that degrades the quality of the generated images, thus we explore different optimization losses for better training of AsymmetricGAN, making image translation with higher consistency and better stability. Extensive experiments on both supervised and unsupervised generative tasks with 8 datasets show that AsymmetricGAN achieves superior model capacity and better generation performance compared with existing GANs. To the best of our knowledge, we are the first to investigate the asymmetric GAN structure on both unsupervised and supervised image translation tasks. 
%   Code is available at \url{https://github.com/Ha0Tang/AsymmetricGAN}.
\end{abstract}

%\begin{CCSXML}
%<ccs2012>
%  <concept_id>10010520.10010553.10010562</concept_id>
%  <concept_desc>Computer systems organization~Embedded systems</concept_desc>
%  <concept_significance>500</concept_significance>
%  <concept_id>10010520.10010575.10010755</concept_id>
%  <concept_desc>Computer systems organization~Redundancy</concept_desc>
%  <concept_significance>300</concept_significance>
%  <concept_id>10010520.10010553.10010554</concept_id>
%  <concept_desc>Computer systems organization~Robotics</concept_desc>
%  <concept_significance>100</concept_significance>
%  <concept_id>10003033.10003083.10003095</concept_id>
%  <concept_desc>Networks~Network reliability</concept_desc>
%  <concept_significance>100</concept_significance>
%</ccs2012>
%\end{CCSXML}

%\ccsdesc[500]{Computer systems organization~Embedded systems}


\maketitle

\section{Introduction}
\label{sec:int}

Recently, Generative Adversarial Networks (GANs)~\cite{goodfellow2014generative} have received considerable attention in computer vision community.
GANs are generative models which are particularly designed for generation tasks.
Recent works have been able to yield promising image translation performance, such as Pix2pix~\cite{isola2017image}, in a supervised setting given carefully annotated image pairs. 
However, pairing the training data is usually difficult and costly. 
% The situation becomes even worse when dealing with tasks such as artistic stylization, since the desired output is very complex, typically requiring artistic authoring. 
To tackle this problem, several GAN approaches, such as CycleGAN~\cite{zhu2017unpaired}, DualGAN~\cite{yi2017dualgan} and ComboGAN~\cite{anoosheh2017combogan}, target to effectively learn a mapping from the source domain to the target domain without paired training data. 
%Some progress has been made by these cross-modal translation frameworks on unpaired image translation tasks.
However, these are not efficient for multi-domain image-to-images translation tasks. 
%For example, for $m$ different domains, BicycleGAN and Pix2pix need the training of $m(m-1)$ models; CycleGAN and DualGAN require $\frac{m(m-1)}{2}$ models; ComboGAN needs to train $m$ models for different $m$ image domains.

%\begin{figure} \small
%	\centering
%	\includegraphics[width=1\linewidth]{framework_comp2.jpg}
%	\caption{An illustration of the unsupervised and supervised frameworks of the proposed AsymmetricGAN (b and d) compared with StarGAN~\cite{choi2017stargan} (a) and GestureGAN~\cite{tang2018gesturegan} (c) for both multi-domain image-to-image translation and hand gesture-to-gesture translation tasks. 
%%		Both StarGAN and GestureGAN share the same generator $G$ for both the image translation and reconstruction tasks.
%%		However, source domain and target domain are asymmetric in complexity, thus both StarGAN and GestureGAN pose the problems of poor generation quality and mapping ambiguity.
%%		While the proposed AsymmetricGAN employs task-specific asymmetric generators, i.e., the translation generator $G^t$ and the reconstruction generator $G^r$, which allows for different network designs and different levels of parameter sharing. The symbols $z_x$ and $z_y$ denote class labels. $l_x$ and $l_y$ represent hand skeletons. $L$ denotes the pixel loss.
%	}
%	\label{fig:motivation}
%	\vspace{-0.4cm}
%\end{figure}

To fix the limitation,
Choi et al. propose StarGAN~\cite{choi2017stargan}, which performs multi-domain image translation using only one generator/discriminator pair and an extra domain classifier~\cite{odena2016conditional}. 
Mathematically, we assume $X$ and $Y$ represent the source and target domains, and $x {\in} X$ and $y {\in} Y$ denote images in domain $X$ and domain $Y$, respectively; we define $z_x$ and $z_y$ denote category labels of domain $X$ and $Y$, respectively.
StarGAN utilizes a symmetric GAN model and uses the same generator $G$ twice to translate $X$ into $Y$ with the target label $z_y$, i.e., $G(x, z_y) {\approx} y$, and then reconstruct the input image $x$ from the translated output $G(x, z_y)$ and the label $z_x$, i.e., $G(G(x, z_y), z_x) {\approx} x$. 
In this way, the generator $G$ shares a common mapping and data structures for two different tasks, i.e., image translation and image reconstruction. 
Moreover, StarGAN cannot handle with some specific image translation tasks such as person image generation~\cite{ma2017pose,siarohin2018deformable} and hand gesture translation~\cite{tang2018gesturegan} since both tasks have infinite image domain $m$ as indicated in~\cite{tang2018gesturegan}.

To solve the limitations, Tang et al. propose GestureGAN~\cite{tang2018gesturegan}, which produces hand gestures with different poses, sizes and locations by using hand skeletons~$l_x$ and~$l_y$.
Note that GestureGAN also uses a symmetric structure, i.e., GestureGAN utilizes the same generator $G$ twice for both image translation and image reconstruction, which can be defined as $G(x, l_y) {\approx} y$ and $G(G(x, l_y), l_x) {\approx} x$, respectively.
In summary, both StarGAN and GestureGAN use a symmetric structure of generators for both image translation and image reconstruction tasks. 
We argue that since each task has unique information and distinct targets, it is harder to optimize the generator and to make it gain a good generalization ability on both tasks.

In this paper, we analyze the limitation of both StarGAN and GestureGAN, and observe that it is their symmetric generator that hinders the improvement of model performance. 
To fix this limitation, we propose a novel asymmetric-structured generator in our Asymmetric Generative Adversarial Network (AsymmetricGAN) for both unsupervised and supervised image translation tasks. 
Unlike StarGAN and GestureGAN, AsymmetricGAN consists of two different asymmetric generators of unequal sizes to adapt to the asymmetric need in both image translation and image construction.

There are three reasons for designing the asymmetric-structured generators.
Firstly, the translation generator $G^t$ transforms images from $X$ to $Y$, and the reconstruction generator $G^r$ uses the translated images from $G^t$ and the original domain guidance $z_x$/$l_x$ to reconstruct the original $x$. 
Generators $G_t$ and $G_r$ cope with different tasks.
Image translation is our main task and image reconstruction is an auxiliary task which aims to provide supervision information to the main task.
Moreover, during testing time, generating a new image from the input image is more difficult than reconstructing the input image since the image reconstruction process has itself as the reference.
Thus we need a stronger generator for image translation and a weaker generator for image reconstruction in our framework.

Secondly, the input data distribution for them is different. 
The inputs of the translation generator $G_t$ are a real image and a target domain guidance. 
The goal of $G_t$ is to generate the target domain image.
While $G_r$ accepts a translated image and an original domain guidance as input, and tries to reconstructs the original input image. 
For generator $G_t$ and $G_r$, the input images are a real image and a generated image respectively, and thus the data distribution is different between them.

Thirdly, because of the complexity difference between the source and target image domains, the complexity inequality in a bidirectional image-to-image translation is significant.
Therefore, it is intuitive to design different network structures for the two different generators, i.e, the powerful network for the complex image translation process and the simple network is used for the simple image reconstruction process.
These two generators are allowed to use different network architecture designs and different levels of parameter sharing strategy according to the diverse difficulty of the tasks. 
By doing so, each generator can have its own network parts which usually helps to learn better each task-specific mapping in a multi-task setting~\cite{ruder2017overview}.

We also investigate how the distinct network designs and different network sharing schemes for the asymmetric generators dealing with different sub-tasks could balance the generation performance and network complexity. 
%A motivation illustration of the proposed AsymmetricGAN compared with the most related two works, i.e., StarGAN and GestureGAN, is presented in Fig.~\ref{fig:motivation}.
Moreover, to avoid the model collapse issue in training AsymmetricGAN for both unsupervised and supervised image-to-image translation tasks, we further explore different objective functions for better optimization. 
1) The color cycle-consistency loss which targets solving the `channel pollution' problem proposed in~\cite{tang2018gesturegan} by separately generating red, green and blue color channels instead of generating all three at one time. 
2) The multi-scale SSIM loss, which preserves the information of luminance, contrast and structure between reconstructed images and input images across different scales.
3) The conditional identity preserving loss, which helps retaining the identity information of the input images. 
These loss functions are jointly embedded in AsymmetricGAN for training and help to generate results with higher consistency and better stability. 

%Extensive experimental results on 8 datasets show that the design of two different generators can alleviate model collapse problem. 
%Qualitative results show the method can generate sharp and reasonable outputs than previous works.
%Moreover, the proposed framework can produces more diverse results on multi-domain style transfer tasks. 
%Finally, we quantitatively compare with existing leading methods in terms of several popular metrics, e.g, IS and FID.
In summary, the main contributions of this paper are:
1) We studies an interesting problem in both unsupervised and supervised image-to-image translation tasks with a novel Asymmetric Generative Adversarial Network (AsymmetricGAN). From architecture to loss functions, we present reasonable and useful solutions.	
2) We propose the asymmetric dual generators, allowing different network structures and different-level parameter sharing, to specifically cope with image translation and image reconstruction tasks, which facilitates obtaining a better generalization ability of the proposed model to improve the generation performance. More importantly, we provide a new direction in designing a good GAN model for both unsupervised and supervised image-to-image translation tasks.
3) We explore jointly utilizing different objectives for a better optimization of the proposed AsymmetricGAN, and thus obtaining both unsupervised and supervised image-to-image translation with higher consistency and better stability.
4) We extensively evaluate AsymmetricGAN on both multi-domain image translation and hand gesture translation tasks with eight different datasets, demonstrating its superiority in model capacity and its better generation performance compared with state-of-the-art methods. 

\hao{A preliminary version of this paper appeared as \cite{tang2018dual}. 
Compared with \cite{tang2018dual}, extensions are made on five aspects.
1) A more detailed analysis is presented in `Introduction' section by giving a deeper analysis on the motivation and the difference from relevant works.
2) A more detailed discussion is presented in `Related Work' section by including recently published works dealing with both unsupervised and supervised image-to-image translation.
3) We extend the model proposed in~\cite{tang2018dual} to a unified GAN framework for handling both unsupervised and supervised image-to-image translation tasks.
4) We present an in-depth description of the proposed approach, providing all the architectural and implementation details of the method, with special emphasis on guaranteeing the reproducibility of the experiments.
5) We extend the experimental evaluation provided in \cite{tang2018dual} in several directions. First, we discuss the generation performance and network complexity of our AsymmetricGAN with different network designs and parameter-sharing strategies on multi-domain image-to-image translation tasks. Second, we conduct extensive experiments on the challenging hand gesture-to-gesture translation task with two different datasets, demonstrating the wide application scope of our GAN framework on both unsupervised and supervised image-to-image translation tasks.}

%This paper is organized as follows.
%Sec.~\ref{sec:rel} surveys the evolution of image-to-image translation related methods.
%Sec.~\ref{sec:for} presents both unsupervised and supervised frameworks of the proposed AsymmetricGAN.
%In Sec.~\ref{sec:imp}, experimental evaluations and detailed discussions on two popular image generation tasks, i.e., multi-domain image-to-image translation and hand gesture-to-gesture translation, are presented.
%Finally, Sec.~\ref{sec:con} concludes this paper.
%%%%%%%%%%%%%%%%%%%%%
\section{Related Work}
\label{sec:rel}

\noindent \textbf{Generative Adversarial Networks (GANs)} \cite{goodfellow2014generative} are generative models, which have achieved promising results on different generative tasks such as image generation.
%and inpainting~\cite{zhang2019gazecorrection,dolhansky2018eye}. 
% The key point for GANs' success lies in the adversarial loss which forces the generated images to be indistinguishable from the real images. 
% However, GANs are difficult to train, since it is hard to keep the balance between the generator and the discriminator, which makes the optimization oscillate and thus leading to a collapse of the generator. 
Moreover, to generate images controlled by users, Mirza et al.~\cite{mirza2014conditional} propose Conditional GANs (CGANs), which use a conditional information to guide the image generation process. 
Extra guidance information can be category labels~\cite{choi2017stargan,xu2019toward}, human skeletons~\cite{siarohin2018deformable} and semantic maps~\cite{wang2018pix2pixHD}. 
In this paper, we mainly focus on image-to-image translation tasks.
%methods of this problem can be divided into two categories, i.e., supervised and unsupervised.

\noindent \textbf{Image-to-Image Translation.} CGANs learn a mapping between image inputs and image outputs using convolutional neural networks. 
For example, Isola et al.~propose Pix2pix~\cite{isola2017image}, which is a conditional framework using a CGAN to learn the mapping function. 
%Based on Pix2pix, Wang et al. present Pix2pixHD~\cite{wang2018pix2pixHD}, which can be used for high-resolution photo-realistic image translation.
Similar ideas have also been applied to many other tasks, e.g., person image generation~\cite{ma2017pose}.
However, all of these models require paired training data, which are usually costly to obtain.
To alleviate the issue of pairing training data, Zhu et al. introduce CycleGAN~\cite{zhu2017unpaired}, which learns the mappings between two unpaired image domains without supervision with the aid of a cycle-consistency loss. 
%Apart from CycleGAN, there are other variants proposed to tackle the same problem such as~\cite{tang2019attention,xu2019toward}.
%For instance, 
%a novel DualGAN mechanism is demonstrated in~\cite{yi2017dualgan}, in which image translators are trained from two unlabeled image sets each representing an image domain. Kim~et al.~\cite{kim2017learning} propose a method based on GANs that learns to discover relations between different domains. 
However, these models are only suitable to cross-domain translation tasks.

\noindent \textbf{Multi-Domain Image-to-Image Translation.} 
There are only very few recent methods attempting to implement multi-domain image-to-image translation in an efficient way. 
Anoosheh et al.~propose ComboGAN~\cite{anoosheh2017combogan}, which only requires to train $m$ generator/discriminator pairs for $m$ different image domains. 
Choi et al.~present StarGAN~\cite{choi2017stargan}, which equips a single symmetric-structured generator to handle the task.
Although the model complexity is low, jointly learning both image translation and image reconstruction with the same generator require the sharing of all parameters, which increases the optimization complexity and reduces the generalization ability, thus leading to unsatisfactory generation performance. 
The proposed method targets at obtaining a good balance between the network capacity and image generation quality. 
Along with this research line, we propose a novel AsymmetricGAN, which 
achieves this target via using two task-specific and asymmetric generators. 
Moreover, we explore various optimization objectives to train better the model to produce more consistent and more stable results.
%%%%%%%%%%%%%%%%%%%%%
\section{AsymmetricGAN}
\label{sec:for}
This paper studies an asymmetric problem in GANs with a new model. 
From architecture to loss functions, we present reasonable and useful solutions.
In this section, we first start with the network architecture of unsupervised AsymmetricGAN  for multi-domain image-to-image translation tasks, and then introduce the network architecture of supervised AsymmetricGAN for hand gesture-to-gesture translation tasks.
Finally, we introduce the network optimization for both frameworks.

\begin{figure} \small
	\centering
	\includegraphics[width=1\linewidth]{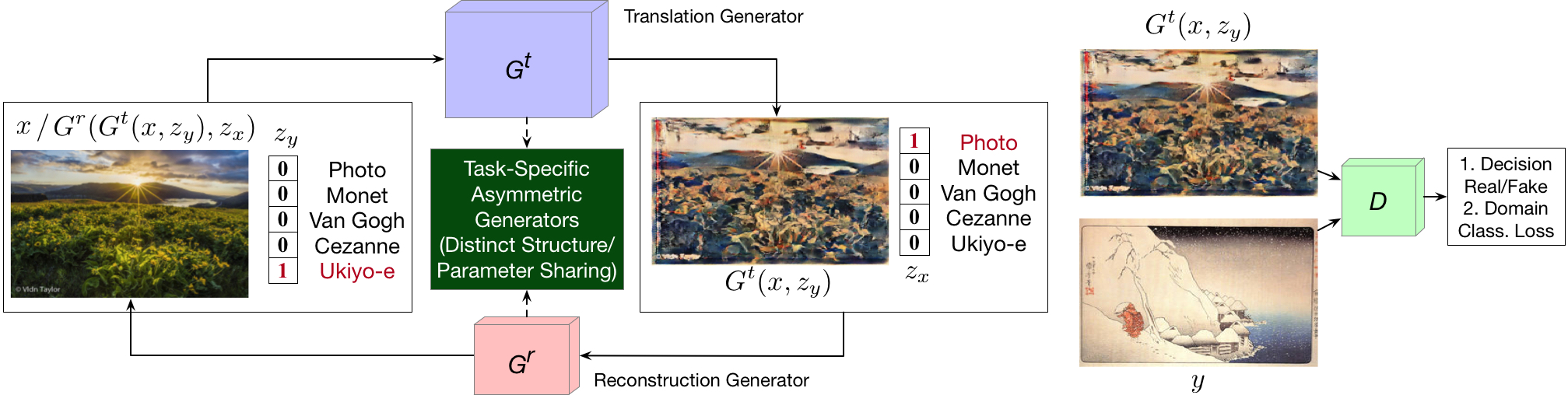}
	\caption{Unsupervised framework of our AsymmetricGAN for multi-domain image-to-image translation. $z_x$ and $z_y$ indicate the category labels of domain $X$ and $Y$, respectively. $G^t$ and $G^r$ are task-specific asymmetric generators. The translation generator $G^t$ translates images from domain $X$ into domain $Y$ and  the reconstruction generator $G^r$ receives the generated image $G^t(x, z_y)$ and the original domain label $z_x$ and tries to recover the input image $x$ during the training stage with the proposed objective functions.}
	\label{fig:mainFramework}
\end{figure}

\subsection{Network Architecture of Unsupervised Framework}
We focus on multi-domain image-to-image translation tasks with unpaired training data. 
The overview of the proposed AsymmetricGAN is illustrated in Fig.~\ref{fig:mainFramework}. 
Existing cross-domain generation models, such as CycleGAN~\cite{zhu2017unpaired}, DiscoGAN \cite{kim2017learning} and DualGAN \cite{yi2017dualgan}, which need to separately train $\frac{m(m-1)}{2}$ models for $m$ different image domains.
However, the proposed AsymmetricGAN is specifically designed for tackling multi-domain image translation problems with significant advantages in the model complexity and in the training overhead, which only needs to train a single model.
To directly compare with StarGAN~\cite{choi2017stargan}, which simply adopts the same generator for both image reconstruction and image translation tasks.
We argue that the training of a single generator model for multiple domains is a challenging problem as mentioned in the introduction section, thus we propose a more effective asymmetric generator network structure and more robust optimization objectives to stabilize the training process. 
In summary, our work focuses on exploring different strategies to improve the optimization of multi-domain translation models, which we aim to give useful insights into the design of more effective multi-domain translation generators. These aspects are not covered and considered in the multi-domain model StarGAN~\cite{choi2017stargan}. 

\begin{table}[!tbp] \small
	\centering
	\caption{Generator network architecture I for multi-domain image-to-image translation.}
	\resizebox{0.5\linewidth}{!}{%
		\begin{tabular}{l|c|c} \toprule		    
			Layer & Input $\rightarrow$  Output Shape      & Layer Information \\ \midrule
			1     & (h, w, 3+z) $\rightarrow$ (h, w, 9)    & Conv-(9, 3, 1, 1), IN, ReLU  \\ 
			2     & (h, w, 9) $\rightarrow$ (h, w, 8)      & Conv-(8, 3, 1, 1), IN, ReLU  \\ 
			3     & (h, w, 8) $\rightarrow$ (h, w, 7)      & Conv-(7, 3, 1, 1), IN, ReLU  \\  
			4     & (h, w, 7) $\rightarrow$ (h, w, 6)      & Conv-(6, 3, 1, 1), IN, ReLU  \\  
			5     & (h, w, 6) $\rightarrow$ (h, w, 5)      & Conv-(5, 3, 1, 1), IN, ReLU  \\  
			6     & (h, w, 5) $\rightarrow$ (h, w, 4)      & Conv-(4, 3, 1, 1), IN, ReLU  \\ 
			7     & (h, w, 4) $\rightarrow$ (h, w, 3)      & Conv-(3, 3, 1, 1), Tanh  \\ 	\bottomrule	
	\end{tabular}}
	\label{tab:gt3}
\end{table}

\begin{table}[!tbp] \small
	\centering
	\caption{Generator network architecture II for multi-domain image-to-image translation.}
	\resizebox{0.6\linewidth}{!}{%
		\begin{tabular}{l|c|c} \toprule		    
			Layer & Input $\rightarrow$  Output Shape      & Layer Information \\ \midrule
			1     & (h, w, 3+z) $\rightarrow$ (h, w, 64)   & Conv-(64, 7, 1, 3), IN, ReLU  \\ 
			2     & (h, w, 64) $\rightarrow$ (h/2, w/2, 128)   & Conv-(128, 4, 2, 1), IN, ReLU  \\ 
			3     & (h/2, w/2, 128) $\rightarrow$ (h/4, w/4, 256)  & Conv-(256, 4, 2, 1), IN, ReLU  \\  
			4     & (h/4, w/4, 256) $\rightarrow$ (h/2, w/2, 128)  & Conv-(128, 4, 2, 1), IN, ReLU  \\  
			5     & (h/2, w/2, 128) $\rightarrow$ (h, w, 64)   & Conv-(64, 4, 2, 1), IN, ReLU  \\  
			6     & (h, w, 64) $\rightarrow$ (h, w, 3)     & Conv-(3, 7, 1, 3), Tanh  \\ 		\bottomrule
	\end{tabular}}
	\label{tab:gt2}
\end{table}

\begin{table}[!tbp] \small
	\centering
	\caption{Generator network architecture III for multi-domain image-to-image translation.}
		\resizebox{0.7\linewidth}{!}{%
		\begin{tabular}{l|c|c} \toprule		    
			Part              & Input $\rightarrow$  Output Shape & Layer Information \\ \midrule
			Down-Sampling     & \tht{c}{(h, w, 3+$z$)$\rightarrow$(h, w, 64) \\ (h, w, 64)$\rightarrow$(h/2, w/2, 128) \\ (h/2, w/2, 128) $\rightarrow$ (h/4, w/4, 256)} & \tht{c}{Conv-(64, 7, 1, 3), IN, ReLU \\ Conv-(128, 4, 2, 1), IN, ReLU \\ Conv-(256, 4, 2, 1), IN, ReLU}   \\ \hline
			Bottleneck        & \tht{c}{(h/4, w/4, 256) $\rightarrow$ (h/4, w/4, 256) \\ (h/4, w/4, 256) $\rightarrow$ (h/4, w/4, 256) \\ (h/4, w/4, 256) $\rightarrow$ (h/4, w/4, 256) \\ (h/4, w/4, 256) $\rightarrow$ (h/4, w/4, 256) \\ (h/4, w/4, 256) $\rightarrow$ (h/4, w/4, 256) \\ (h/4, w/4, 256) $\rightarrow$ (h/4, w/4, 256)} & \tht{c}{RB: Conv-(256, 3, 1, 1), IN, ReLU \\ RB: Conv-(256, 3, 1, 1), IN, ReLU \\ RB: Conv-(256, 3, 1, 1), IN, ReLU \\ RB: Conv-(256, 3, 1, 1), IN, ReLU \\ RB: Conv-(256, 3, 1, 1), IN, ReLU \\ RB: Conv-(256, 3, 1, 1), IN, ReLU}  \\ \hline	
			Up-Sampling       & \tht{c}{(h/4, w/4, 256) $\rightarrow$ (h/2, w/2, 128) \\ (h/2, w/2, 128) $\rightarrow$ (h, w, 64) \\ (h, w, 64)  $\rightarrow$ (h, w, 3) }  & \tht{c}{DConv-(128, 4, 2, 1), IN, ReLU \\ DConv-(64, 4, 2, 1), IN, ReLU \\ Conv-(3, 7, 1, 3), Tanh} \\ 	\bottomrule	
	\end{tabular}}
	\label{tab:gt1}
\end{table}

The proposed AsymmetricGAN consists of an asymmetric dual-generator and a discriminator as shown in Fig.~\ref{fig:mainFramework}. 
The asymmetric dual generators are designed to specifically deal with different tasks in GANs, i.e., the translation and the reconstruction tasks, which have different targets for training the network. 
We can design different network structures for the different generators to make them learn better task-specific objectives, which allows us to share parameters between the generators to further reduce the model capacity since the shallow image representations are shareable for both generators. 
The parameter sharing facilitates the achievement of good balance between the model complexity and the generation quality. 

To achieve this goal, we design three different combination settings (i.e., S1, S2, S3) with three different generator architectures ranging from light-weight to heavy-weight: 
1) Architecture I has the simplest network structure, only consisting of 7 non-linear transformation operations with each using a convolution and a ReLU layer. The number of parameters of this architecture is 2.9K. 
2) Architecture II uses an encoder-decoder network with a symmetric structure, which has 1.3M parameters. 
3) Architecture III employs the same encoder-decoder network as architecture II while adding extra 6 residual blocks. It has the largest network capability (8.4M parameters) in the considered three.
The detailed structures are shown in Tables~\ref{tab:gt3}, \ref{tab:gt2}, \ref{tab:gt1}.

In multi-domain image translation, the final target is to make the network have a good generation ability. 
Thus the translation generator $G^t$ is expected to use a more powerful architecture, while the reconstruction generator $G^r$ can employ a lighter structure. 
We consider the following combinations for the translation and the reconstruction generators: 
1) In S1, $G^t$ uses the generator architecture III, and $G^r$ uses the generator architecture I.
2) In S2, $G^t$ uses the architecture III, and $G^r$ uses the generator architecture II.
3) In S3, $G^t$ and $G^r$ use the same generator architecture III.
Note that the proposed model generalizes the state-of-the-art model StarGAN~\cite{choi2017stargan}. 
When the parameters are fully shared with the usage of the same network structure for both generators, our framework becomes a StarGAN.

For each combination, the translation generator $G^t$ is learned to translate an input image $x$ into an output image $y$ which is conditioned on the target domain label $z_y$, this process can be expressed as~$G^t(x, z_y){\rightarrow} y$. 
Then the reconstruction generator $G^r$ receives the translated image $G^t(x, z_y)$ and the original domain label $z_x$ as input, and learns to recover the input image $x$, this process can be formulated as $G^r(G^t(x, z_y), z_x){\rightarrow} x$. 
We represent the class labels $z_x$ and $z_y$ using a one-hot vector, and then the vector is passed through a linear layer to obtain a label embedding with 64 dimensions. 
This embedding is replicated to form feature maps that are further concatenated with the image feature maps for follow-up convolution operations with residual blocks and several deconvolution layers to obtain the target images.
%The asymmetric generators are task-specific generators which allow for different network designs and different levels of parameter sharing for learning better the generators. 
The discriminator $D$ tries to distinguish between the real image $y$ and the generated image $G^t(x, z_y)$, and also to classify the translated image $G^t(x, z_y)$ to the corresponding domain label $z_y$ via the proposed domain classification loss.  
We adopt PatchGAN~\cite{zhu2017unpaired,choi2017stargan} as the architecture of discriminator~$D$.

\begin{figure} \small
	\centering
	\includegraphics[width=1\linewidth]{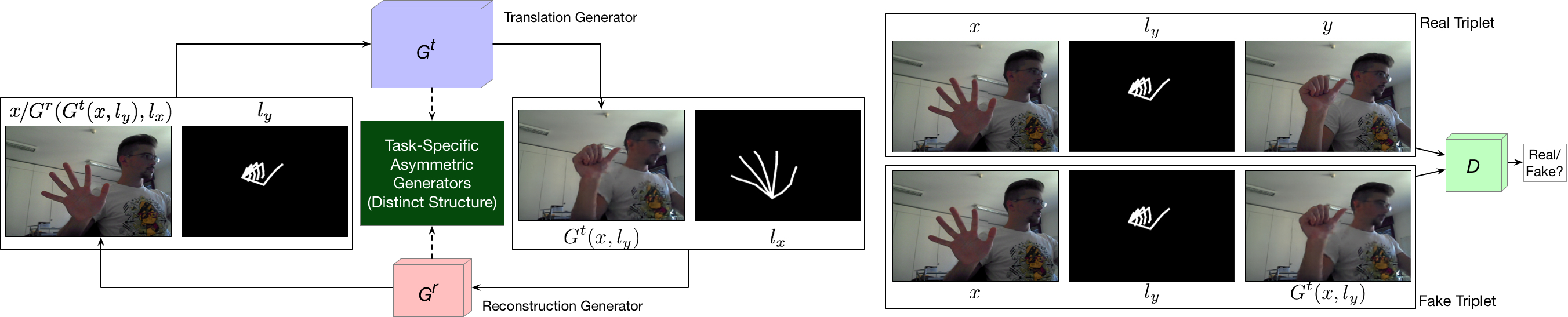}
	\caption{Supervised framework of our AsymmetricGAN for hand gesture-to-gesture translation. $l_x$ and $l_y$ indicate the hand skeletons of images $x$ and $y$, respectively.  $G^t$ and $G^r$ are task-specific asymmetric generators. The translation generator $G^t$ converts images from domain $X$ into domain $Y$ and the construction generator $G^r$ receives the generated image $G^t(x, l_y)$ and the original hand skeleton $l_x$ and attempts to reconstruct the original image $x$ during the optimization with the proposed different objective losses. We have two cycles, i.e., $x {\mapsto} y' {\mapsto} \widehat{x} {\approx} x$ and  $y {\mapsto} x' {\mapsto} \widehat{y} {\approx} y$, but we only show one here, i.e., $x {\mapsto} y' {\mapsto} \widehat{x} {\approx} x$.}
	\label{fig:mainFramework_sup}
\end{figure}

\subsection{Network Architecture of Supervised Framework}
In this part, we start to introduce the supervised framework of the proposed AsymmetricGAN for hand gesture-to-gesture.
The framework of the proposed AsymmetricGAN is shown in Fig.~\ref{fig:mainFramework_sup}, which consists of asymmetric dual generators (i.e., $G^t$ and $G^r$) and a discriminator~$D$.
Specifically, we concatenate the input image $x$ and the target hand skeleton $l_y$, and input them into the translation generator $G^t$ and synthesize the target image $y'{=}G^t(x, l_y)$. 
Different from GestureGAN~\cite{tang2018gesturegan}, which adopts the same generator to reconstruct the original input image, we propose an asymmetric  reconstruction generator $G^r$ to benefit more from the image translation process. 
The conditional hand skeleton $l_x$ together with the generated image $y'$ are input into the reconstruction generator $G^r$, and reconstruct the original input image $\widehat{x}$. 
We formalize the process as $\widehat{x}{=}G^r(y',l_x){=}G^r(G^t(x,l_y), l_x)$.
Then the optimization objective is to make $\widehat{x}$ as close as possible to $x$.

We adopt the architecture from~\cite{johnson2016perceptual} as our generators $G^t$ and $G^r$.
Since our main task is the image translation, it means that $G^t$ should be more powerful than $G^r$. 
Therefore we use a deeper network for $G^t$ and a shallow network for $G^r$ to adapt to the asymmetric translations.
Specifically, we use nine residual blocks for both generators.
However, the filters in first convolutional layer of $G^t$ and $G^r$ are 64 and 4, respectively.
The network of $G^t$ consists of:
$c7s1\_64, d128, d256, R256, R256, R256, R256, R256, R256, R256, R256, R256$,\\$u128, u64, c7s1\_3$, where $c7s1\_k$ denote a $7 {\times}7$ Convolution-InstanceNormReLU layer with $k$ filters and stride 1; 
$dk$ denotes a $3 {\times} 3$ Convolution-InstanceNorm-ReLU layer with $k$ filters and stride 2;
$Rk$ denotes a residual block that contains two $3 {\times} 3$ convolutional layers with the same number of filters on both layer;
$uk$ denotes a $3 {\times} 3$ fractional-strided-ConvolutionInstanceNorm-ReLU layer with $k$ filters and stride $1/2$.
The network of $G^r$ consists of: \\
$c7s1\_4, d8, d16, R16, R16, R16, R16, R16, R16, R16, R16, R16, u8, u4, c7s1\_3$.\\
Thus, the total parameter of $G^t$ and $G^r$ is 11.388 M and 0.046~M, respectively.
For the discriminator $D$, we adopt $70 {\times} 70$ PatchGAN proposed in~\cite{isola2017image}.

\subsection{Network Optimization}
In this section, we first introduce the same optimization functions between unsupervised and supervised frameworks of AsymmetricGAN, then we separately introduce different optimization functions of the two frameworks.
These optimization losses are jointly embedded into the proposed AsymmetricGAN during the training stage.

\noindent\textbf{Optimization Functions of Both Frameworks.} The same optimization functions between unsupervised and supervised frameworks are the color cycle-consistency loss and conditional identity preserving loss.

\noindent\textit{Color Cycle-Consistency Loss.}
The cycle-consistency loss can be regarded as `pseudo' pairs in training data even though we do not have corresponding samples in the target domain. This loss function can be defined as,
\begin{equation} 
\begin{aligned}
& \mathcal{L}_{cyc}(x)  = \mathbb{E}_{x\sim{p_{\rm data}}(x)} \left[|| G^r(G^t(x, g_y), g_x)-x||_1 \right],
\end{aligned}
\label{equ:cycleganloss}
\end{equation}
where ($g_x$, $g_y$) can be one of ($z_x$, $z_y$) and ($l_x$, $l_y$) according to the tasks, i.e, ($z_x$, $z_y$) for unsupervised tasks and ($l_x$, $l_y$) for supervised tasks.
%We follow \cite{isola2017image} and use $L1$ distance for the reconstruction loss since $L1$ encourages less blurring than $L2$.
The goal of this loss is to make the reconstructed image $G^r(G^t(x, g_y), g_x)$ as close as possible to the input image~$x$.
However, the generation of a whole image at one time makes the different color channels influence each other, thus leading to artifacts in the generation results \cite{tang2018gesturegan}. 
To overcome this limitation, we propose a novel color cycle-consistency loss, which constructs the consistence loss for each channel, separately. 
Thus, this loss can be expressed as,
\begin{equation} 
\begin{aligned}
\mathcal{L}_{colorcyc}=\sum_{i\in\{r,g,b\}}\mathcal{L}_{cyc}(x^i),
\end{aligned}
\label{equ:colorcycleganloss}
\end{equation}
where ${x^b, x^g, x^r}$ denote the blue, \hao{green} and red channels of the image $x$.
We calculate the pixel loss for the red, green, blue channel separately  and then sum up these three color losses as the final loss. 
%In this way, the generator can be enforced to generate each channel independently to avoid the `channel pollution' problem. 
%Note that different from PG$^2$~\cite{ma2017pose}, DPIG~\cite{ma2017disentangled} and PoseGAN~\cite{siarohin2018deformable}, which only use the target skeletons to guide the image generation process, the proposed AsymmetricGAN not only uses the target skeletons to guide the image translation process, but also uses them to guide the image reconstruction process. In this way, the cycle consistency can further be guaranteed.

\noindent\textit{Conditional Identity Preserving Loss.}
To reinforce the identity information during the translation process, we use a conditional identity preserving loss~\cite{zhu2017unpaired}.
This loss encourages the mapping to preserve identity information such as color information between the input images and the output images,
%\begin{equation}
%\begin{aligned}
%& \mathcal{L}_{id}(G^t, G^r, z_x)= \mathbb{E}_{x\sim{p_{\rm data}}(x)}\left[|| G^r(x,z_x)-x||_1\right].
%\end{aligned}
%\label{eq:id}
%\end{equation}
\begin{equation}
\begin{aligned}
\mathcal{L}_{id}=  \mathbb{E}_{x\sim{p_{\rm data}}(x)} \left[|| G^t(x, g_x)-x||_1 \right]  
+ \mathbb{E}_{y\sim{p_{\rm data}}(y)} \left[|| G^t(y, g_y)-y ||_1 \right]. 
\end{aligned}
\label{eq:id2}
\end{equation}  
where ($g_x$, $g_y$) can be one of ($z_x$, $z_y$) and ($l_x$, $l_y$) according to the tasks.
%We adopt the $L1$ loss to minimize the difference between the input image ($x$ or $y$) and the self-guided result ($G^t(x, g_x)$ or $G^t(y, g_y)$).
%Thus, the generator preserves the identity via the back-propagation of this loss.
%Without this loss, the generators are free to change the tint of input images when there is no need to.

\noindent\textbf{Optimization Functions of Unsupervised Framework.}
Other optimization objects for the unsupervised framework are the conditional least square loss, the multi-scale SSIM loss and the domain classification loss.

\noindent\textit{Conditional Least Square Loss.}  We use a least square loss \cite{mao2017least,zhu2017unpaired} to stabilize our model during the training stage, which is more stable than the negative log likelihood objective. 
%$ \mathcal{L}_{cgan}(G^t, D_s, z_y) {=} \mathbb{E}_{y\sim{p_{\rm data}}(y)}\left[ \log D_s(y)\right] + \mathbb{E}_{x\sim{p_{\rm data}}(x)}\left[\log (1 - D_s(G^t(x, z_y)))\right]$.
%and is converging faster than Wasserstein GAN~\cite{arjovsky2017wasserstein}. 
This loss can be expressed as:
\begin{equation} 
\begin{aligned}
\mathcal{L}_{lsgan} =  \mathbb{E}_{y\sim{p_{\rm data}(y)}} \left[(D_s(y)-1)^2\right]  
+  \mathbb{E}_{x\sim{p_{\rm data}}(x)} \left[D_s( G^t(x, z_y))^2\right],
\end{aligned}
\label{equ:legan}
\end{equation}
where $z_y$ are the category label of $y$, $D_s$ is the probability distribution over sources produced by discriminator~$D$. 
The target of $G^t$ is to generate an image $G^t(x,z_y)$ that is expected to be similar to the images from domain $Y$, while $D$ aims to distinguish the generated image $G^t(x,z_y)$ from the real~$y$.

\noindent\textit{Multi-Scale SSIM Loss.}
The structural similarity index (SSIM) has been originally proposed in~\cite{wang2004image} to measure the similarity of two images. 
We introduce it here to help to preserve the information of luminance, contrast and structure across scales. 
For the reconstructed image $\widehat{x}{=}G^r(G^t(x,z_y),z_x)$ and the input image $x$, the SSIM loss is written as:
\begin{equation} 
\begin{aligned}
& \mathcal{L}_{ssim}(\widehat{x},x)=\left[l(\widehat{x},x)\right]^\alpha \left[c(\widehat{x}, x)\right]^\beta \left[s(\widehat{x},x) \right]^\gamma,
\end{aligned}
\end{equation}
where 
$l(\widehat{x},x){=} \frac{2\mu_{\widehat{x}}\mu_x + C_1}{\mu_{\widehat{x}}^2 +\mu_x^2 +C_1}$, 
$c(\widehat{x},x){=} \frac{2\sigma_{\widehat{x}}\sigma_x + C_2}{\sigma_{\widehat{x}}^2 + \sigma_x^2 + C_2}$ and 
$s(\widehat{x},x){=} \frac{\sigma_{{\widehat{x}}x}+C_3}{\sigma_{\widehat{x}}\sigma_x+C_3}$.
These three terms compare the luminance, contrast and structure information between $\widehat{x}$ and $x$. 
$\alpha$, $\beta$ and $\gamma$ are hyper-parameters to control the relative weight of $l(\widehat{x},x)$, $c(\widehat{x},x)$ and $s(\widehat{x},x)$, respectively; $\mu_{\widehat{x}}$ and $\mu_x$ are the means of $\widehat{x}$ and $x$; $\sigma_{\widehat{x}}$ and $\sigma_x$ are the standard deviations of $\widehat{x}$ and $x$; $\sigma_{\widehat{x}x}$ is the covariance of $\widehat{x}$ and $x$; $C_1$, $C_2$ and $C_3$ are predefined parameters. To make the model benefit from multi-scale deep information, we refer to a multi-scale implementation of SSIM~\cite{wang2003multiscale} which constrains SSIM over $M$ scales,
\begin{equation} 
\begin{aligned}
& \mathcal{L}_{msssim}(\widehat{x},x) = \left[l_M(\widehat{x},x)\right]^{\alpha_M} \prod\limits_{j=1}^M \left[ c_j(\widehat{x},x)\right]^{\beta_j} \left[s_j(\widehat{x},x)\right]^{\gamma_j}.
\end{aligned}
\label{equ:ssim}
\end{equation}
%Through using this loss, the luminance, contrast and structure information of the input images is expected to be preserved.

\noindent\textit{Domain Classification Loss.}
To perform multi-domain image translation with a single discriminator, we employ an auxiliary classifier~\cite{odena2016conditional,choi2017stargan} on the top of the discriminator, and impose the domain classification loss when updating both the generator and discriminator.
\begin{equation} 
\begin{aligned}
\mathcal{L}_{class}=\mathbb{E}_{x\sim{p_{\rm data}(x)}} \left\{ -\left[\log D_c(z_x|x) + \log D_c(z_y|G^t(x, z_y))\right] \right\},
\end{aligned}
\label{equ:class}
\end{equation}
where $D_c(z_x|x)$ represents the probability distribution over the domain labels given by discriminator $D$. 
$D$ learns to classify~$x$ to its corresponding domain $z_x$. 
$D_c(z_y|G^t(x, z_y)$ denotes the domain classification for fake images. We minimize the domain classification loss to produce the image $G^t(x, z_y)$ that can be classified to the corresponding domain $z_y$.

\noindent\textit{Full Objective of Unsupervised Framework.} Given the loss functions presented above, the complete optimization objective of Unsupervised AsymmetricGAN for multi-domain image-to-image translation can be written as:
\begin{equation}
\begin{aligned}
\mathcal{L} = \mathcal{L}_{lsgan} + \lambda_{c}\mathcal{L}_{class} + \lambda_{cyc} \mathcal{L}_{colorcyc} + \lambda_{m} \mathcal{L}_{msssim} + \lambda_{id} \mathcal{L}_{id},
\end{aligned}
\label{eqn:allloss}
\end{equation}
where $\lambda_{c}$, $\lambda_{cyc}$, $\lambda_{m}$ and $\lambda_{id}$ are parameters controlling the relative importance of the corresponding objective. All objectives are jointly optimized in an end-to-end fashion.
We follow previous works \cite{zhu2017unpaired,choi2017stargan} and set $\lambda_{c}{=}1$, $\lambda_{cyc}{=}10$, $\lambda_{m}{=}1$, $\lambda_{id}{=}0.5$ in our experiments.

\noindent\textbf{Optimization Functions of Supervised Framework.}
Other optimization objects for the supervised framework are the conditional adversarial loss, the improved pixel loss, the perceptual loss, and the total variation loss.

\noindent\textit{Conditional Adversarial Loss.} 
%The goal of vanilla GAN loss is to train the generator $G$ which learns the mapping from random noise $z$ to the image $y$.
%The mapping $G(z) {\rightarrow} y$ can be learned through the following function,
%\begin{equation}
%\begin{aligned}
%\mathcal{L}_{gan}(G^t, D) = & \mathbb{E}_{y\sim{p_{\rm data}}(y)}\left[ \log D(y)\right] + \\
%& \mathbb{E}_{z\sim{p_{\rm data}}(z)}\left[\log (1 - D(G^t(z)))\right].
%\end{aligned}
%\end{equation}
%Base on this, 
The goal of CGANs try to learn the mapping from a conditional image $x$ to the target image $y$.
The generator $G^t$ tries to generate image $y'{=}G^t(x)$ which cannot be distinguished from the real image $y$, while the discriminator $D$ tries to detect the fake images produced by~$G^t$.
%Thus, the objective function of CGANs can be expressed as,
\begin{equation}
\begin{aligned}
\mathcal{L}_{cgan}(G^t, D) = \mathbb{E}_{y\sim{p_{\rm data}}(y)}\left[ \log D(x, y)\right] 
+ \mathbb{E}_{x\sim{p_{\rm data}}(x)}\left[\log (1 - D(x, G^t(x, l_y)))\right],
\end{aligned}
\label{eq:cgan}
\end{equation}
where $D$ tries to distinguish the fake image pair $(x, G^t(x, l_y))$ from the real image pair $(x, y)$.
To jointly learn the images and the hand skeletons, we make a modification base on Eq.~\eqref{eq:cgan},
\begin{equation}
\begin{aligned}
\mathcal{L}_{cgan}(G^t, D) = \mathbb{E}_{y\sim{p_{\rm data}}(y)}\left[ \log D(x, l_y, y)\right] + \mathbb{E}_{x\sim{p_{\rm data}}(x)}\left[\log (1 - D(x, l_y, G^t(x, l_y)))\right],
\end{aligned}
\label{eq:cgan2}
\end{equation}
where $D$ tries to distinguish the fake image triplet $(x, l_y, G^t(x, l_y))$ from the real image triplet $(x, l_y, y)$.
In this way, $D$ takes consideration of both images and hand skeletons during optimization.

\noindent\textit{Improved Pixel Loss.} 
%Since the paired training data are available for this task, 
We also adopt an improved pixel loss between generated images and ground truths, i.e., channel-wise color loss,  to reduce the `channel pollution' issue~\cite{tang2018gesturegan}.
The loss can be expressed as,
\begin{equation}
\begin{aligned}
\mathcal{L}_{color}
= \sum_{i\in\{r,g,b\}} \mathbb{E}_{x\sim{p_{\rm data}}(x)}\left[|| G^t(x^i, l_y) -y^i||_1\right].
\end{aligned}
\label{eqn:color2}
\end{equation}
where $y^r$, $y^g$ and $y^b$ denote the red, green and blue color channels of image $y$. 
%The intuition is that the generation of a three-channel image is much more complex than the generation of a one-channel image.
By calculating the loss of each channel independently, the error from each channel will not influence other channels.

\noindent\textit{Perceptual Loss and Total Variation Loss.}
We also use a perceptual loss and a total variation loss between the generated image $y'$ and the real image $y$ to better optimize our model. 
Both losses have been shown to be useful in Pix2pixHD~\cite{wang2018pix2pixHD} and SelectionGAN~\cite{tang2019multi}, respectively.

\noindent\textit{Full Objective of Supervised Framework.} The final objective of AsymmetricGAN for hand gesture-to-gesture translation can be expressed as, 
\begin{equation}
\begin{aligned}
\mathcal{L} = 
& \mathcal{L}_{cgan} + \lambda_{c}\mathcal{L}_{color} +
\lambda_{cyc} \mathcal{L}_{colorcyc}
+ \lambda_{id} \mathcal{L}_{id}  + \lambda_{vgg} \mathcal{L}_{vgg}+ \lambda_{tv} \mathcal{L}_{tv}, 
\end{aligned}
\label{eqn:allloss2}
\end{equation}
where hyper-parameters $\lambda_{c}$, $\lambda_{cyc}$, $\lambda_{id}$, $\lambda_{vgg}$ and $\lambda_{tv}$ are controlling the relative importance of each loss.
%All objectives are jointly optimized in an end-to-end fashion. 
In our experiments, we follow previous works \cite{isola2017image,tang2018gesturegan} and empirically set $\lambda_{c}{=}800$, $\lambda_{cyc}{=}0.1$, $\lambda_{id}{=}0.01$, $\lambda_{vgg}{=}1000$ and $\lambda_{tv}{=}1\mathrm{e}{-}6$.

%\noindent\textbf{Network Training.} 
%We employ the Adam optimizer~\cite{kingma2014adam} with $\beta_1{=} 0.5$ and $\beta_2{=} 0.999$ to optimize the whole model. 
%We sequentially update $G^t$ and $G^r$ after $D$ updates at each iteration. 
%The batch size is set to 4 for both datasets and all the models are trained with 20 epochs. 
%Moreover, we follow~\cite{ma2017pose,tang2018gesturegan} and  employ OpenPose~\cite{simon2017hand} to extract hand skeletons as training data.
\begin{table} \small
	\centering
	\caption{Description of datasets used in multi-domain image-to-image translation.}
	\resizebox{0.8\linewidth}{!}{%
		\begin{tabular}{l|c|c|c|c|c|c|c|c} \toprule		    
			Dataset                                              & Type                    & \#Domain & \#Mapping & Resolution            & Unpaired/Paired & \#Train   & \#Test   & \#Total \\ \midrule
			Facades~\cite{tylevcek2013spatial}   & Architectures      & 2               & 2                 & 256$\times$256    & Paired               & 800           & 212          & 1,012 \\ 
			AR Face~\cite{martinez1998ar}       & Faces                  & 4              & 12                   & 768$\times$576    & Paired                & 920           & 100          & 1,020 \\ 
			Bu3dfe~\cite{yin20063d}        & Faces                  & 7               & 42                   & 512$\times$512    & Paired               & 2,520         & 280          & 2,800 \\ 
			Alps~\cite{anoosheh2017combogan} & Natural Seasons & 4                & 12                   & -                           & Unpaired          & 6,053         & 400          & 6,453 \\ 
			RaFD~\cite{langner2010presentation} & Faces                  & 8               & 56                  & 1024$\times$681   & Unpaired          &  5,360       & 2,680       & 8,040 \\ 
			Collection  Style~\cite{zhu2017unpaired} & Painting Style      & 5               & 20	                & 256$\times$256     & Unpaired          & 7,837         & 1,593        & 9,430 \\		\bottomrule
	\end{tabular}}
	\label{tab:dataset}
\end{table}

\section{Experiments}
\label{sec:imp}

We conduct experiments on two challenging tasks to evaluate the effectiveness of our AsymmetricGAN, i.e., multi-domain image-to-image translation and hand gesture-to-gesture translation.
%We employ the Adam optimizer~\cite{kingma2014adam} with $\beta_1{=}0.5$ and $\beta_2{=}0.999$ to optimize the whole model. 
%We sequentially update the translation generator $G^t$ and the reconstruction generator $G^r$ after the discriminator $D$ updates at each iteration. 
%The proposed AsymmetricGAN is implemented using deep learning framework PyTorch. 
%Experiments are conducted on NVIDIA TITAN Xp GPUs. 

\subsection{Multi-Domain Image-to-Image Translation}

\subsubsection{Experimental Setup} 

\noindent\textbf{Datasets and Parameter Settings.}
%We employ 6 datasets to validate our AsymmetricGAN on multi-domain image-to-image translation tasks, i.e., Facades~\cite{tylevcek2013spatial}, AR Face~\cite{martinez1998ar}, Alps Season~\cite{anoosheh2017combogan}, Bu3dfe~\cite{yin20063d}, RaFD~\cite{langner2010presentation}, and Collection Style~\cite{zhu2017unpaired}.
%Table~\ref{tab:dataset} shows the details of these datasets.
We employ 6 datasets to validate our AsymmetricGAN on multi-domain image-to-image translation tasks.
Table~\ref{tab:dataset} shows the details of these datasets.
%For reducing model oscillation, we adopt a cache of generated images to update the discriminator as in~\cite{shrivastava2017learning}. 
%In the experiments, we set the number of image buffer to 50. 
The batch size is set to 1 for all the experiments and all the models are trained with 200 epochs. 
%We keep the same learning rate for the first 100 epochs and linearly decay the rate to zero during the next 100 epochs. 
%The initial learning rate for the Adam optimizer is 0.0002. 
%Weights were initialized from a Gaussian distribution with mean 0 and standard deviation 0.02.

\noindent\textbf{Competing Baselines.} 
We employ several models as our baselines, i.e., CycleGAN~\cite{zhu2017unpaired}, DualGAN~\cite{yi2017dualgan}, ComboGAN~\cite{anoosheh2017combogan}, DistanceGAN~\cite{benaim2017one}, Dist.+Cycle~\cite{benaim2017one}, Self Dist.~\cite{benaim2017one},  BicycleGAN~\cite{zhu2017toward}, Pix2pix~\cite{isola2017image}, and StarGAN~\cite{choi2017stargan} . 
%We train the models multiple times for every pair of two different image domains except for ComboGAN~\cite{anoosheh2017combogan}, which only needs to train $m$ models for $m$ different domains.
%We also adopt StarGAN~\cite{choi2017stargan} as a baseline which performs multi-domain image translation using one generator/discriminator pair. 
%The fully supervised Pix2pix and BicycleGAN are trained with paired data when available, the other baselines and the proposed AsymmetricGAN are trained with unpaired data.
%Note that since BicycleGAN can produce several different outputs with a single input image, then we randomly pick one output from them for comparison.
%For fair comparisons, results of all baselines are produced by using the authors' publicly available codes with the same training strategy as our approach.

\begin{figure}[!tbp] \small
	\centering
	\includegraphics[width=0.8\linewidth]{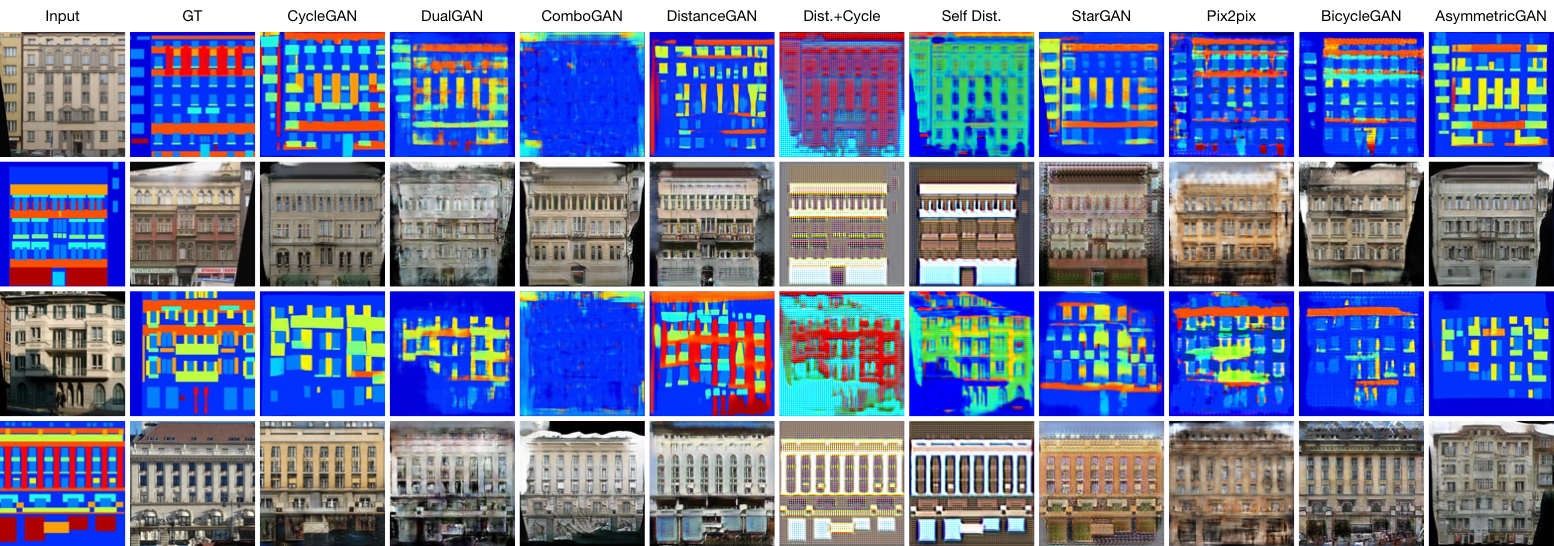}
	\caption{Different methods for label$\leftrightarrow$photo translation on Facades. 
	\hao{From left to right: Input, Ground Truth (GT), CycleGAN~\cite{zhu2017unpaired}, DualGAN~\cite{yi2017dualgan}, ComboGAN~\cite{anoosheh2017combogan}, DistanceGAN~\cite{benaim2017one}, DistanceGAN+Cycle Loss~\cite{benaim2017one}, DistanceGAN+Self Distance~\cite{benaim2017one}, StarGAN~\cite{choi2017stargan}, Pix2pix~\cite{isola2017image}, BicycleGAN~\cite{zhu2017toward}, and AsymmetricGAN (Ours).}
	} 
	\label{fig:comparison_facades}
\end{figure}

\begin{figure}[!tbp] \small
	\centering
	\includegraphics[width=0.8\linewidth]{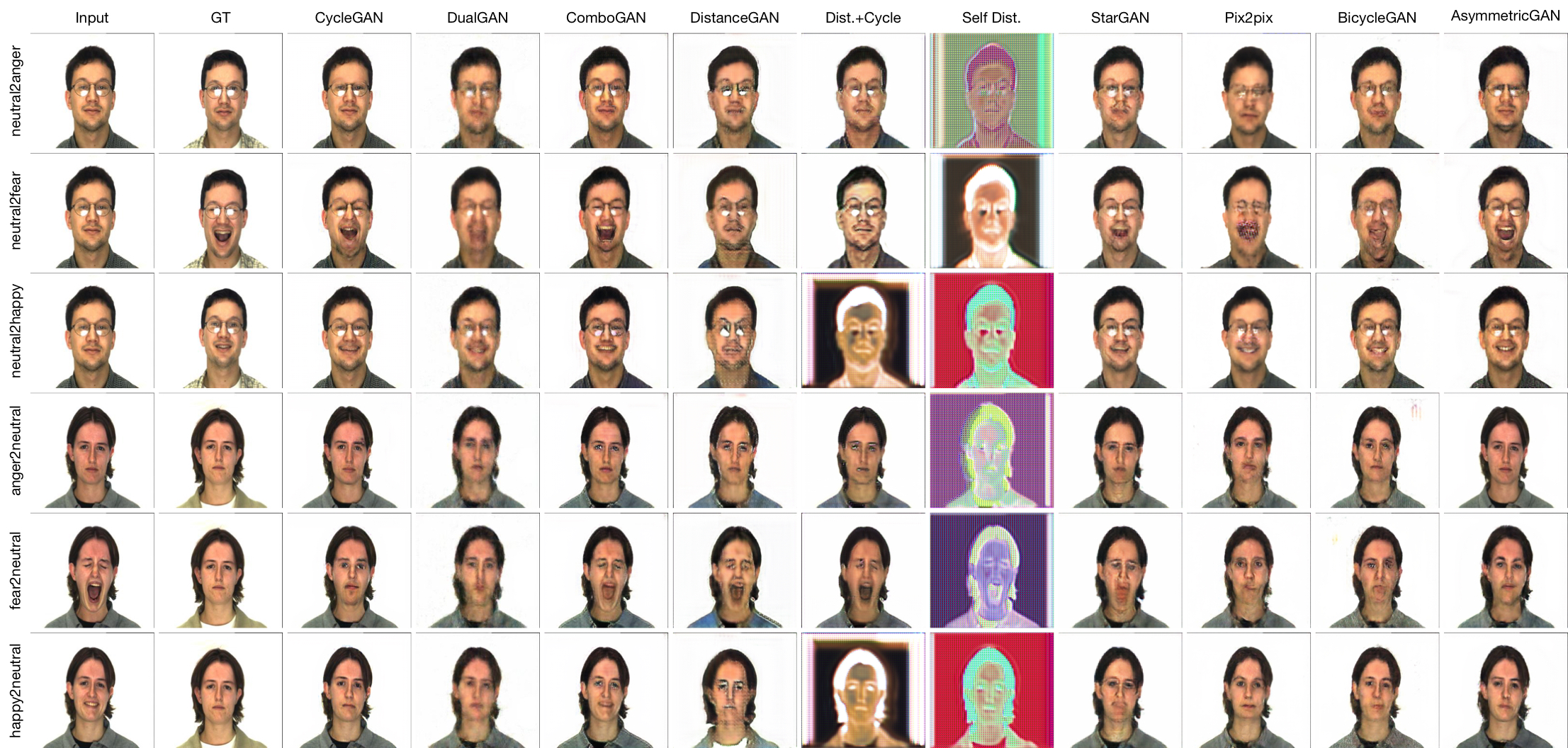}
	\caption{Different methods for multi-domain facial expression translation on AR Face. 
		\hao{From left to right: Input, Ground Truth (GT), CycleGAN~\cite{zhu2017unpaired}, DualGAN~\cite{yi2017dualgan}, ComboGAN~\cite{anoosheh2017combogan}, DistanceGAN~\cite{benaim2017one}, DistanceGAN+Cycle Loss~\cite{benaim2017one}, DistanceGAN+Self Distance~\cite{benaim2017one}, StarGAN~\cite{choi2017stargan}, Pix2pix~\cite{isola2017image}, BicycleGAN~\cite{zhu2017toward}, and AsymmetricGAN (Ours).}
	}
	\label{fig:comparison_ar}
\end{figure}

\subsubsection{Comparison Against State-of-the-Arts}
%The proposed AsymmetricGAN is evaluated on four different tasks, i.e., label$\leftrightarrow$photo translation, facial expression-to-expression translation, season translation and painting style transfer. 
%We will describe the comparison with the state-of-the-art methods in the following.

\begin{figure}[!tbp] \small
	\centering
	\includegraphics[width=0.8\linewidth]{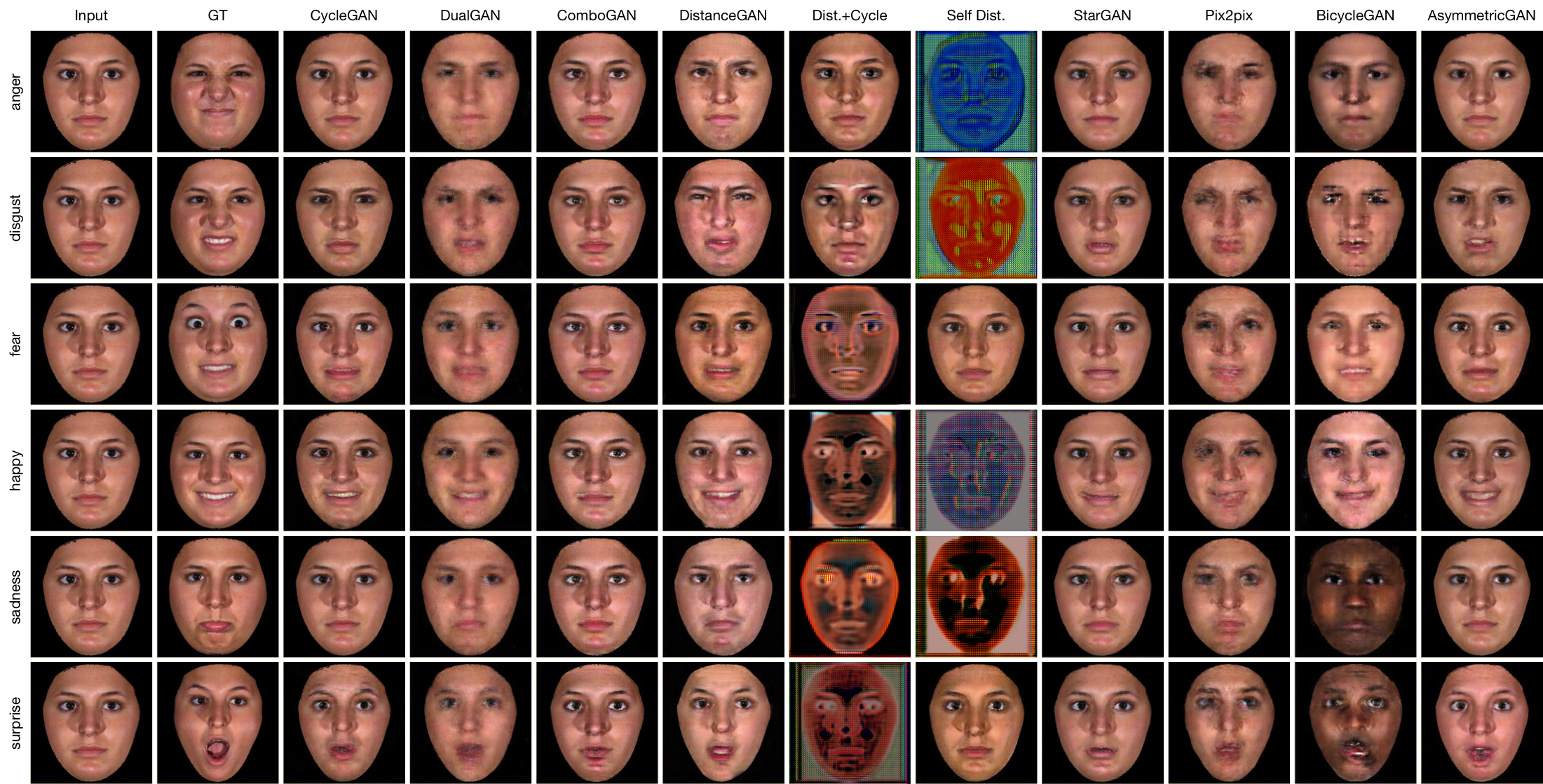}
	\caption{Different methods for multi-domain facial expression translation on  Bu3dfe. 
		\hao{From left to right: Input, Ground Truth (GT), CycleGAN~\cite{zhu2017unpaired}, DualGAN~\cite{yi2017dualgan}, ComboGAN~\cite{anoosheh2017combogan}, DistanceGAN~\cite{benaim2017one}, DistanceGAN+Cycle Loss~\cite{benaim2017one}, DistanceGAN+Self Distance~\cite{benaim2017one}, StarGAN~\cite{choi2017stargan}, Pix2pix~\cite{isola2017image}, BicycleGAN~\cite{zhu2017toward}, and AsymmetricGAN (Ours).}
	}
	\label{fig:comparison_bu3dfe}
\end{figure}

\noindent\textbf{Task 1: Label$\leftrightarrow$Photo Translation.}
We use Facades to perform label$\leftrightarrow$photo translation, which we aim to show that AsymmetricGAN is also applicable on the translation on two domains only and could produce competitive performance.
Results are shown in Fig.~\ref{fig:comparison_facades}.
We can see that Dist.+Cycle, Self Dist., ComboGAN fail to generate reasonable results on the photo$\rightarrow$label task. 
For the opposite mapping, i.e., label$\rightarrow$photo, Dist.+Cycle, Self Dist., DualGAN, StarGAN and Pix2pix suffer from model collapse, leading reasonable but blurry results. 
However, the proposed AsymmetricGAN achieves compelling results in both directions compared with baselines. 

\begin{figure}[!tbp] \small
	\centering
	\includegraphics[width=0.8\linewidth]{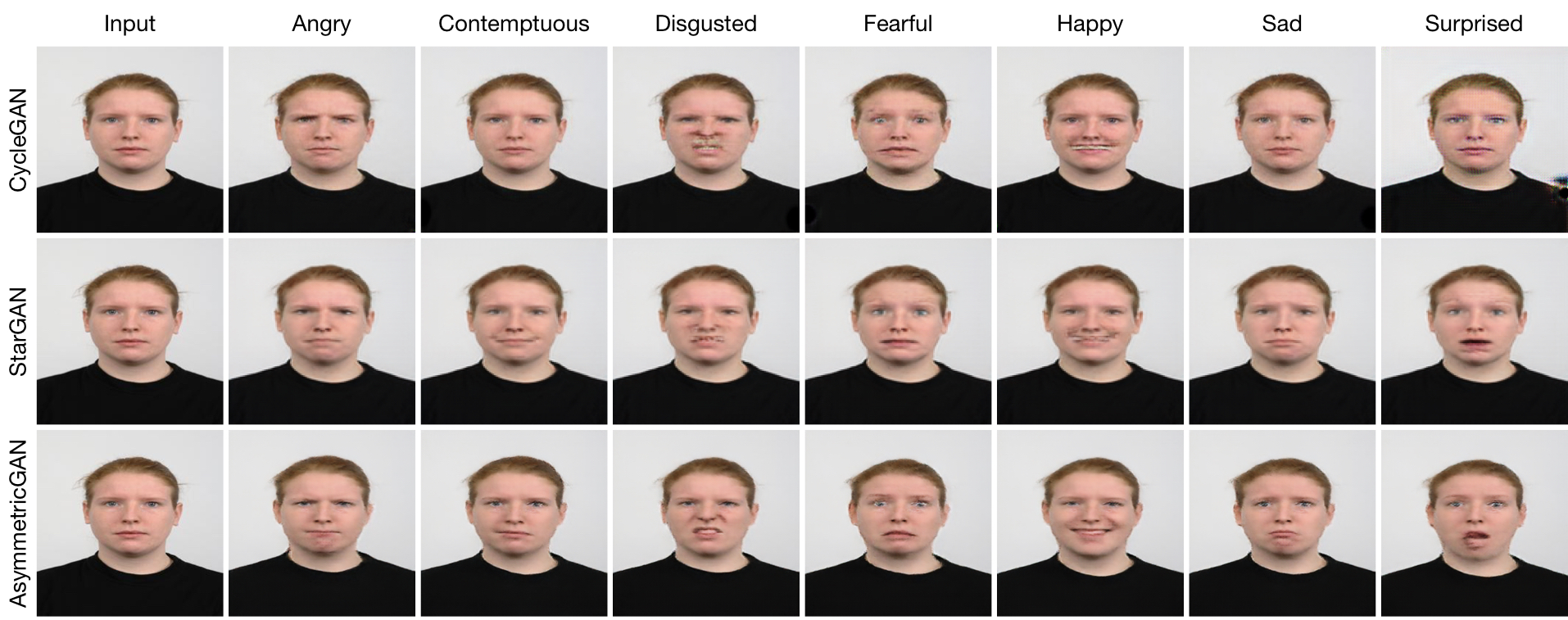}
	\caption{Different methods for multi-domain facial expression translation on RaFD. \hao{From top to bottom: CycleGAN~\cite{zhu2017unpaired}, StarGAN~\cite{choi2017stargan}, and AsymmetricGAN (Ours).}
	}
	\label{fig:comparison_rafd}
\end{figure}

\begin{figure}[!tbp] \small
	\centering	
	\includegraphics[width=0.8\linewidth]{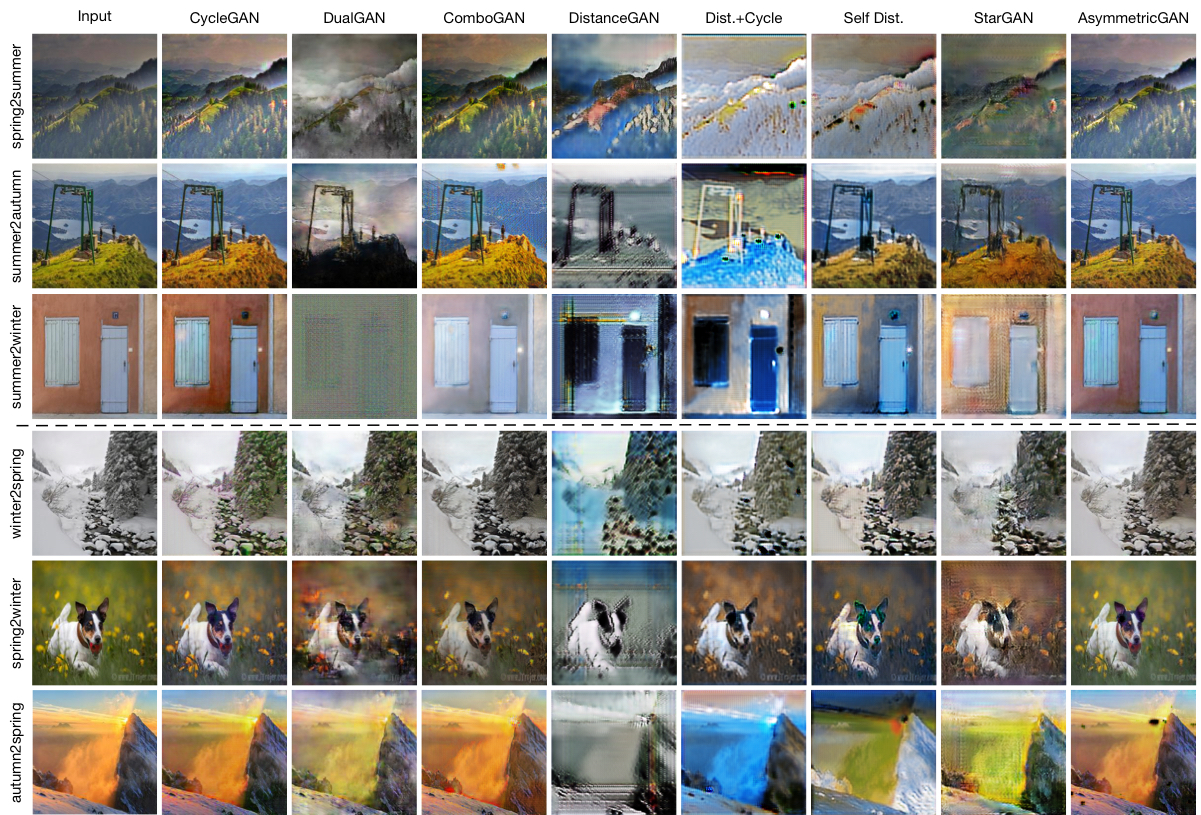}
	\caption{
		Different methods for multi-domain season translation on Alps. 
		\hao{From left to right: Input, CycleGAN~\cite{zhu2017unpaired}, DualGAN~\cite{yi2017dualgan}, ComboGAN~\cite{anoosheh2017combogan}, DistanceGAN~\cite{benaim2017one}, DistanceGAN+Cycle Loss~\cite{benaim2017one}, DistanceGAN+Self Distance~\cite{benaim2017one}, StarGAN~\cite{choi2017stargan}, and AsymmetricGAN (Ours). }
%		Three failure cases are shown in the last three rows.
	}
	\label{fig:comparison_alps}
\end{figure}

\noindent\textbf{Task 2: Facial Expression Synthesis.}
We employ three face datasets, i.e., AR Face, Bu3dfe and RaFD, to evaluate facial expression synthesis tasks. 
%Note that for AR dataset, we not only show the translation results of the neutral expression to other non-neutral expressions as in~\cite{choi2017stargan}, but also present the opposite mappings, i.e.,~from non-neutral expressions to neutral expression. 
%For Bu3dfe dataset, we only show the translation results from neutral expression to other non-neutral expressions as in~\cite{choi2017stargan}. 
Results of AR Face are shown in Fig.~\ref{fig:comparison_ar}, we can see that Dist.+Cycle and Self Dist. fail to produce faces similar to the target domain. 
DualGAN generates reasonable but blurry faces. 
DistanceGAN, StarGAN, BicycleGAN and Pix2pix produce much sharper results, but still contain some artifacts in the translated faces, e.g., twisted mouths on StarGAN, Pix2pix and BicycleGAN in the `neutral2fear' direction. 
ComboGAN, CycleGAN and the proposed AsymmetricGAN work better than other baselines. 
Similar results can be seen on Bu3dfe as shown in Fig.~\ref{fig:comparison_bu3dfe}.
We also present results on RaFD compared with the most related two works, i.e., CycleGAN and StarGAN, in Fig.~\ref{fig:comparison_rafd}. 
We see that our method achieves visually better results than both.

\begin{figure}[!tbp] \small
	\centering
	\includegraphics[width=0.8\linewidth]{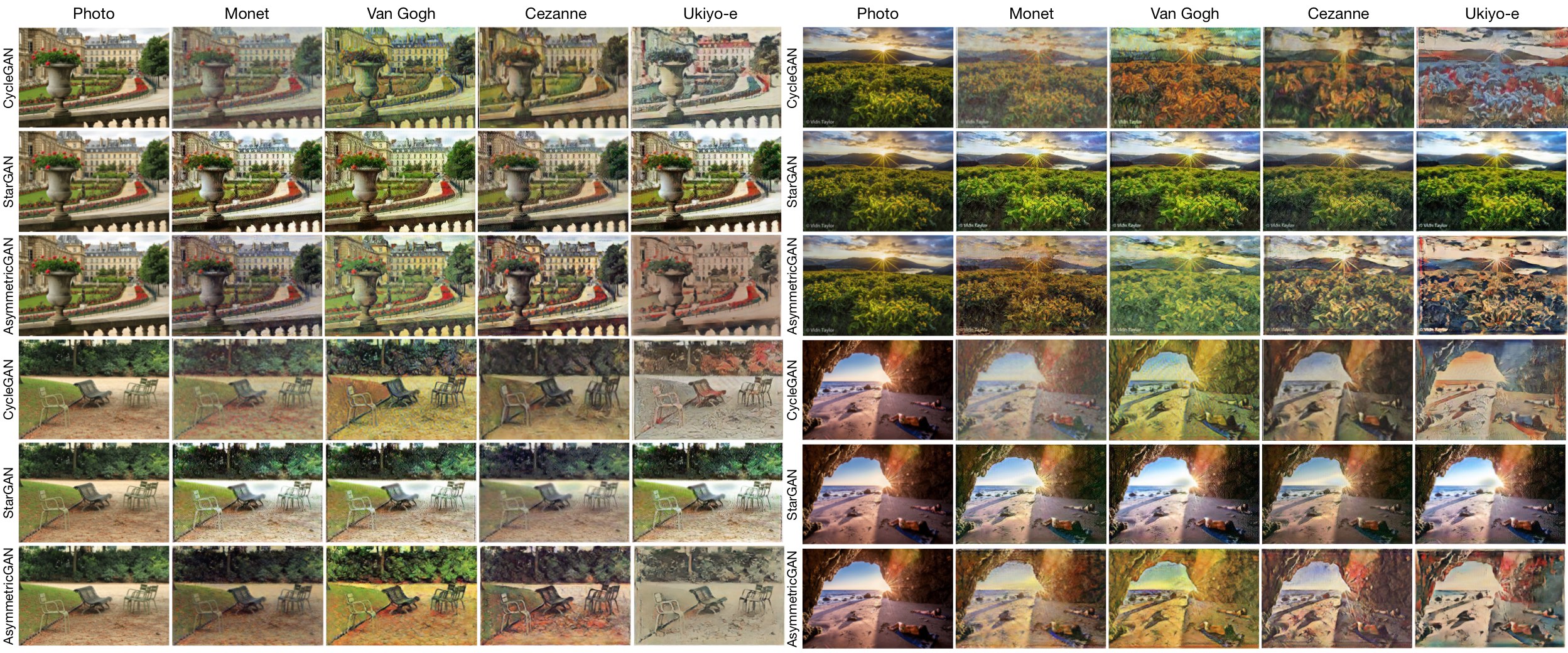}
	\caption{Different methods for multi-domain painting style transfer on Collection Style. 
	\hao{From top to bottom: CycleGAN~\cite{zhu2017unpaired}, StarGAN~\cite{choi2017stargan}, and AsymmetricGAN (ours).}
		Our method achieves significantly better results than StarGAN.}
	\label{fig:style_results}
\end{figure}

\noindent\textbf{Task 3: Season Translation.}
Fig.~\ref{fig:comparison_alps} shows the season translation results. 
%Note that we did not show both Pix2pix and BicycleGAN results on Alps since this dataset does not contain ground-truth images to train these two models. 
Clearly, DistanceGAN, Dist.+Cycle, Self Dist., DualGAN fail to produce reasonable results.
StarGAN can generate reasonable but blurry results, and there are some visual artifacts in the translated results.
ComboGAN, CycleGAN  and the proposed AsymmetricGAN are able to produce better results than other methods. 
However, ComboGAN yields some visual artifacts in some cases, such as in the `summer2autumn' direction.
We also show three failure case of the proposed method on this dataset as shown in Fig.~\ref{fig:comparison_alps}. 
Our method generates images similar to the input domain, while existing methods such as CycleGAN and DualGAN generate visually better results compared with the proposed AsymmetricGAN in the `winter2spring', `spring2winter' and `autumn2spring' directions.
However, both DualGAN and CycleGAN require to train 12 generators for this task on the dataset, while the proposed AsymmetricGAN only needs to train 2 generators, and thus our model complexity is significantly lower.

\noindent\textbf{Task 4: Painting Style Transfer.}
Comparison results on painting style transfer tasks compared with the most related two methods, i.e., CycleGAN and StarGAN, are shown in Fig.~\ref{fig:style_results}. 
We see that StarGAN generates less diverse generations crossing different styles compared with CycleGAN and AsymmetricGAN. 
The proposed AsymmetricGAN has comparable performance with CycleGAN, requiring only one single model for all the styles, and thus the network complexity is remarkably lower compared with CycleGAN which trains an individual model for each pair of styles.

\begin{table}[!tbp] \small
	\centering
	\caption{AMT of multi-domain image translation on Facades, AR Face, Alps and Bu3dfe. 
	}
	\resizebox{0.8\linewidth}{!}{%
		\begin{tabular}{lccccc} \toprule
			AMT $\uparrow$   & label$\rightarrow$photo  & photo$\rightarrow$label   & AR Face           & Alps               & Bu3dfe                     \\ \midrule
			
			CycleGAN \cite{zhu2017unpaired}  & 8.8$\pm$1.5        & 4.8$\pm$0.8         & \textbf{24.3$\pm$1.7}& 39.6$\pm$1.4 & 16.9$\pm$1.2         \\ 
			DualGAN \cite{yi2017dualgan}    & 0.6$\pm$0.2        & 0.8$\pm$0.3        & 1.9$\pm$0.6        & 18.2$\pm$1.8 & 3.2$\pm$0.4          \\ 
			ComboGAN \cite{anoosheh2017combogan} & 4.1$\pm$0.5   & 0.2$\pm$0.1   & 4.7$\pm$0.9        & 34.3$\pm$2.2 & \textbf{25.3$\pm$1.6}\\ 
			DistanceGAN \cite{benaim2017one}   & 5.7$\pm$1.1        & 1.2$\pm$0.5         & 2.7$\pm$0.7        & 4.4$\pm$0.3  & 6.5$\pm$0.7          \\  
			Dist.+Cycle \cite{benaim2017one}  & 0.3$\pm$0.2       & 0.2$\pm$0.1         & 1.3$\pm$0.5        & 3.8$\pm$0.6  & 0.3$\pm$0.1          \\  
			Self Dist. \cite{benaim2017one}  & 0.3$\pm$0.1       & 0.1$\pm$0.1         & 0.1$\pm$0.1        & 5.7$\pm$0.5  & 1.1$\pm$0.3          \\  
			StarGAN \cite{choi2017stargan}  & 3.5$\pm$0.7       & 1.3$\pm$0.3         & 4.1$\pm$1.3    & 8.6$\pm$0.7  & 9.3$\pm$0.9          \\ 
			Pix2pix \cite{isola2017image}   & 4.6$\pm$0.5       & 1.5$\pm$0.4         & 2.8$\pm$0.6        & -                        & 3.6$\pm$0.5          \\ 
			BicycleGAN \cite{zhu2017toward}   & 5.4$\pm$0.6   & 1.1$\pm$0.3        & 2.1$\pm$0.5        & -                  & 2.7$\pm$0.4          \\ \hline 		
			Ours, Fully-Sharing          & 4.6$\pm$0.9       & 2.4$\pm$0.4         & 6.8$\pm$0.6        & 15.4$\pm$1.9 & 13.1$\pm$1.3         \\ 
			Ours, Partially-Sharing   & 8.2$\pm$1.2        & 3.6$\pm$0.7        & 16.8$\pm$1.2        & 36.7$\pm$2.3 & 18.9$\pm$1.1        \\
			Ours, No-Sharing    & \textbf{10.3$\pm$1.6}& \textbf{5.6$\pm$0.9}& 22.8$\pm$1.9  & \textbf{47.7$\pm$2.8} & 23.6$\pm$1.7  \\	\bottomrule	
	\end{tabular}}
	\label{tab:result_amt}
\end{table}

\noindent\textbf{Quantitative Comparison on All Tasks.} 
%We also provide quantitative performance on different metrics, i.e.,
%AMT perceptual studies~\cite{zhu2017unpaired,isola2017image}, IS~\cite{salimans2016improved}, FID~\cite{heusel2017gans} and Classification Accuracy (CA)~\cite{choi2017stargan}. 
We follow both CycleGAN and StarGAN, and use the same perceptual study protocol to evaluate the generated images, which is a perceptual metric to assess the realism from a holistic level.  
As we can see from Tables~\ref{tab:result_amt}, \ref{tab:result_paint} and \ref{tab:rafd}, AsymmetricGAN achieves very competitive results compared with baselines. 
Moreover, we observe that AsymmetricGAN significantly outperforms StarGAN trained using one generator on most of the metrics and on all the datasets.
We also note that supervised Pix2pix shows worse results than unpaired methods in Table~\ref{tab:result_amt}, which can be also observed in~\cite{yi2017dualgan}.

We then adopt IS~\cite{salimans2016improved} to measure the quality of synthesized images. 
IS tries to capture two properties of generated images, i.e., image quality and diversity.
Results compared with the most related two works (i.e., CycleGAN and StarGAN) are shown in Table \ref{tab:rafd}. 
We see that our method achieves better IS than both CycleGAN and StarGAN.
Moreover, we employ FID~\cite{heusel2017gans} to evaluate the performance on both RaFD and painting style datasets.
Results compared with CycleGAN and StarGAN are shown in Tables \ref{tab:result_paint} and \ref{tab:rafd}, we observe that AsymmetricGAN achieves the best results compared with StarGAN and CycleGAN.
We finally compute Classification Accuracy (CA) on the generated images as in~\cite{choi2017stargan}. 
%We train different classifiers on the AR Face, Alps, Bu3dfe, Collection Style datasets, respectively.
%For each dataset, we take the real image as training data and the generated images of different models as testing data. 
%The intuition behind this setting is that if the generated images are realistic and follow the distribution of the images in the target domain, the classifiers trained on real images will be able to classify the generated image correctly. 
%For AR Face, Alps and Collection Style, we report top 1 classification accuracy.
%For Bu3dfe, we present both top 1 and top 5 classification accuracies. 
Table \ref{tab:result_paint}  shows the results on style transfer task. 
We see that AsymmetricGAN significantly outperforms both CycleGAN and StarGAN.
	
\begin{table}[!tbp] \small
	\centering
	\caption{AMT, FID and CA of multi-domain image translation on Collection Style. 
	}
	\resizebox{0.45\linewidth}{!}{%
		\begin{tabular}{lccc} \toprule
			Model                                               &  AMT  $\uparrow$            &  FID $\downarrow$  & CA (\%) $\uparrow$\\ \midrule
			CycleGAN~\cite{zhu2017unpaired}   &  16.8$\pm$1.9              &  47.4823             & 73.72 \\ 
			StarGAN~\cite{choi2017stargan}      &  13.9$\pm$1.4             &  58.1562              & 44.63 \\ 
			AsymmetricGAN &  \textbf{19.8$\pm$2.4}    &  \textbf{43.7473} &  \textbf{78.84}  \\ \hline
			Real Data   &  -        &  -                         & 91.34 \\	\bottomrule
	\end{tabular}}
	\label{tab:result_paint}
\end{table}

\begin{table}[!tbp] \small
	\centering
	\caption{AMT, IS and FID of multi-domain image translation on RaFD. 
	}
	\resizebox{.45\linewidth}{!}{%
		\begin{tabular}{lccc} \toprule
			Model                       			           & AMT $\uparrow$  & IS $\uparrow$ & FID $\downarrow$\\ \midrule		
			CycleGAN~\cite{zhu2017unpaired}  & 19.5         &  1.6942     &   52.8230 \\ 
			StarGAN~\cite{choi2017stargan}     & 24.7         &   1.6695     & 51.6929 \\ 
			AsymmetricGAN                              & \textbf{29.1}& \textbf{1.7187}   & \textbf{51.2765}\\ 	\bottomrule
	\end{tabular}}
	\label{tab:rafd}
\end{table}

\subsubsection{Model Analysis} 
%We investigate four aspects of AsymmetricGAN on multi-domain image-to-image translation.

\noindent\textbf{Importance of Distinct Network Designs for Different Generators.}
We first report the results of the running-time for training one epoch, the total number of generator parameters and the quantitative performance on Bu3dfe. 
The network architectures of different generator combinations (i.e., S1, S2, S3) are described in Sec.~\ref{sec:for} and results are compared with the most related model StarGAN as shown in Table~\ref{tab:ff}. 
Note that we also tried increasing the depth and channel number of the generator in StarGAN in our preliminary experiments, but did not observe improved performance. 
Thus, we intuitively replaced the symmetric generator in StarGAN with the proposed asymmetric dual-generator, and the performance has improved significantly.
Specifically, we observe in Table~\ref{tab:ff} that AsymmetricGAN achieves much better performance than StarGAN on all metrics when we consider only a light-weight generator structure for the reconstruction generator (S1). 
By so doing, the number of parameters for ours is only 2.9k more than StarGAN, while the performance is significantly boosted, which shows that the distinct network designs for different generators are very important for learning better both the generators, demonstrating our initial motivation. 

\begin{table}[!tbp] \small
	\centering
	\caption{Comparison results with different generator settings of AsymmetricGAN on Bu3dfe. StarGAN++ uses the same optimization objectives as our method.}
	\resizebox{0.8\linewidth}{!}{
		\begin{tabular}{lccccc} \toprule		    
			Model & \#Time  & \#Parameters   & AMT $\uparrow$  & IS $\uparrow$ & CA (\%) $\uparrow$ \\ \midrule
			StarGAN~\cite{choi2017stargan} & 2m23s  & 8.4M                 &  9.3$\pm$0.9  & 1.5640 & @1:52.704, @5:94.898 \\ 
			StarGAN++~\cite{choi2017stargan} & 2m26s & 8.4M & 12.4$\pm$1.2 & 1.6532 & @1:53.765, @5:95.125 \\ \hline
			S1: $G^t$ (Architecture III), $G^r$ (Architecture I)             & 2m27s & 8.4M+2.9K        &  18.9$\pm$1.4 & 1.8790 & @1:55.575, @5:96.014 \\
			S2: $G^t$ (Architecture III), $G^r$ (Architecture II)            & 2m29s & 8.4M+1.3M       &  20.1$\pm$1.4 & 1.9293 & @1:56.173, @5:97.112 \\
			S3: $G^t$ (Architecture III), $G^r$ (Architecture III)           & 2m33s  & 8.4M+8.4M       &  23.6$\pm$1.7 & 1.8714 & @1:55.625, @5:96.250 \\  \bottomrule	
	\end{tabular}}
	\label{tab:ff}
\end{table}

\begin{table}[!tbp] \small
	\centering
	\caption{Ablation study of AsymmetricGAN on Facades, AR Face and Bu3dfe for multi-domain image translation. 
		All: AsymmetricGAN, I: Identity preserving loss, S: multi-scale SSIM loss, C: Color cycle-consistency loss, D: Double discriminators strategy.}
	\resizebox{0.8\linewidth}{!}{%
		\begin{tabular}{l|c|c|cc|ccc} \toprule
			\multirow{2}{*}{Model}   & Label$\rightarrow$Photo        & Photo$\rightarrow$Label        & \multicolumn{2}{c|}{AR Face}                       & \multicolumn{3}{c}{Bu3dfe}   \\ \cline{2-8}
			& AMT $\uparrow$& AMT $\uparrow$& AMT $\uparrow$& CA (\%) $\uparrow$           & AMT $\uparrow$ & CA (\%) $\uparrow$  & FID $\downarrow$ \\ \midrule	
			All                      & \textbf{10.3$\pm$1.6}    & \textbf{5.6$\pm$0.9}              & \textbf{22.8$\pm$1.9 }             & @1:\textbf{29.667}  & \textbf{23.6$\pm$1.7}               & @1:\textbf{55.625}, @5:\textbf{96.250}  & \textbf{33.28} \\ 
			All - I                  & 2.6$\pm$0.4              & 4.2$\pm$1.1              & 4.7$\pm$0.8              & @1:29.333  & 16.3$\pm$1.1               & @1:53.739, @5:95.625 & 43.89 \\ 
			All - S - C              & 4.4$\pm$0.6             & 4.8$\pm$1.3           & 8.7$\pm$0.6              & @1:28.000  & 14.4$\pm$1.2               & @1:42.500, @5:95.417 & 45.50 \\ 
			All - S - C - I          & 2.2$\pm$0.3              & 3.9$\pm$0.8          & 2.1$\pm$0.4              & @1:24.667  & 13.6$\pm$1.2               & @1:41.458, @5:95.208 & 47.62 \\ 
			All - D                  & 9.0$\pm$1.5             & 5.3$\pm$1.1              & 21.7$\pm$1.7             & @1:28.367  & 22.3$\pm$1.6               & @1:53.375, @5:95.292 & 35.67 \\ 
			All - D - S              & 3.3$\pm$0.7             & 4.5$\pm$1.1            & 14.7$\pm$1.7              & @1:27.333  & 20.1$\pm$1.4               & @1:42.917, @5:91.250  & 38.09 \\ 
			All - D - C              & 8.7$\pm$1.3             & 5.1$\pm$0.9            & 19.4$\pm$1.5              & @1:28.000  & 21.6$\pm$1.4               & @1:45.833, @5:93.875 & 36.12 \\ \bottomrule		
	\end{tabular}}
	\label{tab:Ablation}
\end{table}

Moreover, to study the effectiveness of the proposed asymmetric structure and remove the impact of the proposed optimization functions, we conduct experiments with StarGAN using the same optimization objectives as our method. 
Results are shown in Table \ref{tab:ff}, we observe that the proposed method still achieves much better results than StarGAN++, which further validates our design motivation.

\noindent\textbf{Generation Performance v.s. Network Complexity.}
Through a comparison of the performance among the setting S1, S2, S3 in Table~\ref{tab:ff}, we also observe that using a more complex generator indeed improves the generation performance, while the network capacity is consequently increased. Specifically, from S2 to S3, the number of parameter changes from 8.4M+1.3M to 8.4M+8.4M. Although the parameters remarkably increase, the generation performance has slight improvement (IS and CA metrics are even worse), meaning that the balance between the network complexity and the generation performance should be also considered in designing a good GAN.

\begin{table}[!tbp] \small
	\centering
	\caption{Comparison of the overall model capacity of different models with the number of image domain $m{=}7$ for multi-domain image-to-image translation tasks.}
	\resizebox{0.45\linewidth}{!}{%
		\begin{tabular}{lc|c} \toprule		    
			Model & \#Models   & \#Parameters\\ \midrule
			\tht{l}{Pix2pix \cite{isola2017image} \\ BicycleGAN \cite{zhu2017toward} } & $A_m^2=m(m-1)$ & \tht{l}{57.2M$\times$42 \\ 64.3M$\times$42} \\ \hline
			\tht{l}{CycleGAN \cite{zhu2017unpaired} \\ DiscoGAN \cite{kim2017learning} \\ DualGAN \cite{yi2017dualgan}\\ DistanceGAN \cite{benaim2017one}} & $C_m^2=\frac{m(m-1)}{2}$ & \tht{l}{
				52.6M$\times$21 \\ 16.6M$\times$21 \\ 178.7M$\times$21 \\ 
				52.6M$\times$21} \\ \hline
			ComboGAN \cite{anoosheh2017combogan} & $m$ & 14.4M$\times$7\\ \hline
			StarGAN \cite{choi2017stargan} & 1 & 53.2M$\times$1 \\ \hline
			Ours, Fully-Sharing    & 1 & 53.2M$\times$1  \\
			Ours, Partial-Sharing  & 1 & 53.8M$\times$1 \\
			Ours, No-Sharing	    & 1 & 61.6M$\times$1 \\		\bottomrule
		\end{tabular}}
		\label{tab:computational}
	\end{table}

\noindent \textbf{Efficiency.}
Table \ref{tab:ff} provides the running-time of one epoch on different methods.
We see that our proposed method is only slightly slower than StarGAN++ \cite{choi2017stargan}.
Specifically,  StarGAN++ needs 2m26s to finish one epoch, while the proposed baseline S2 only needs 2m29s for training one epoch.
However, our baseline S2 achieves remarkably better results than StarGAN on all three evaluation metrics.
Moreover,  since the introduction of the proposed asymmetric structure, we observe that our model converges more easily and quickly, thus making image translation with higher consistency and better stability.
Therefore, in order to balance performance and efficiency, we adopt baseline S2 in our unsupervised image-to-image translation experiments.

\noindent\textbf{Model Component Analysis.}
We conduct an ablation study of the proposed AsymmetricGAN on several datasets, i.e., Facades, AR Face and Bu3dfe. 
We report the generation performance without using the conditional identity preserving loss (I), multi-scale SSIM loss (S), color cycle-consistency loss (C) and double discriminators strategy (D), respectively. 
We also employ two different discriminators as in~\cite{nguyen2017dual,tang2018gesturegan} to further improve our generation performance. 
In order to investigate the parameter-sharing strategy of the asymmetric generator, we perform experiments on different schemes including: 
1) Fully-sharing, i.e.,~the two generators share the same parameters. 2) Partially-sharing, i.e.,~only the encoder part shares the same parameters.
3) No-sharing, i.e.,~two independent generators. 
The basic generator structure follows StarGAN~\cite{choi2017stargan}. 
Quantitative results of both AMT score and CA are reported in Table~\ref{tab:Ablation}.
We observe that without using double discriminators slightly degrades performance, meaning that the proposed model can achieve good results trained using the proposed asymmetric dual generators and one discriminator. However, removing the conditional identity preserving loss, multi-scale SSIM loss and color cycle-consistency loss substantially degrades the performance, meaning that the proposed joint optimization objectives are particularly important to stabilize the training process and thus produce much better generation performance. 
Results of different parameter-sharing strategies are shown in Tables~\ref{tab:result_amt} and \ref{tab:computational}, we observe that different-level parameter sharing influences both the generation performance and the model capacity, demonstrating our initial motivation.

\noindent\textbf{Overall Model Capacity Analysis.}
We also compare the overall model capacity with several state-of-the-art methods.
The number of trained models and the number of model parameters on Bu3dfe for $m$ domains are presented in Table~\ref{tab:computational}.
%We note that BicycleGAN and Pix2pix are supervised models, thus they need to train $A_m^2$ models for $m$ domains.
%CycleGAN, DiscoGAN, DualGAN, DistanceGAN are unsupervised methods, and they require $C_m^2$ models to learn $m$ domains, but each model of them contains two generators and two discriminators.
%ComboGAN needs only $m$ models to learn all the mappings of $m$ domains, while 
We note that both StarGAN and AsymmetricGAN only need to train one model to learn all the mappings of $m$ domains. 
We also report the number of parameters on Bu3dfe in Table~\ref{tab:computational}.
This dataset contains 7 different facial expressions, which means $m{=}7$. 
%Note that DualGAN employs the fully connected layers in its generators, which brings a significantly larger number of parameters. 
%CycleGAN and DistanceGAN adopt the same generator and discriminator architectures, which means they have the same number of parameters. 
The proposed AsymmetricGAN uses fewer parameters compared with the other baselines except for StarGAN, but we achieve significantly better generation results in most metrics as shown in Tables \ref{tab:result_amt}, \ref{tab:result_paint} and  \ref{tab:rafd}. 
When we adopt a parameter-sharing strategy, our generation performance is only slightly lower (but still outperforming StarGAN) while the number of parameters is comparable with StarGAN.   

\begin{table}[!tbp] \small
	\centering
	\caption{Comparison between SymmetricGAN and AsymmetricGAN for hand gesture-to-gesture translation tasks on NTU Hand Digit.
	}
	\resizebox{0.6\linewidth}{!}{%
		\begin{tabular}{lccccc} \toprule
			Model                 & PSNR $\uparrow$ & AMT $\uparrow$  & FID $\downarrow$ & FRD $\downarrow$ & \#Parameters \\ \midrule			
			SymmetricGAN   &32.5740 & 27.9 & 6.8711 & 1.7519  & 11.388M*2\\
			AsymmetricGAN &\textbf{32.6686}  & \textbf{29.7} & \textbf{6.7132} & \textbf{1.7341} & \textbf{11.388M+0.046M} \\ 	\bottomrule	
	\end{tabular}}
	\label{tab:gesture_abla}
\end{table}

\begin{table}[!tbp] \small
	\centering
	\caption{Comparison with different models for hand gesture translation.
	}
	\resizebox{0.8\linewidth}{!}{
		\begin{tabular}{lcccccccc} \toprule
			\multirow{2}{*}{Model} & \multicolumn{4}{c}{NTU Hand Digit}& \multicolumn{4}{c}{Senz3D}   \\ \cmidrule(lr){2-5}\cmidrule(lr){6-9}
			& PSNR $\uparrow$ & AMT $\uparrow$   & FID $\downarrow$ & FRD $\downarrow$ & PSNR $\uparrow$ & AMT $\uparrow$ & FID $\downarrow$ & FRD $\downarrow$  \\ \midrule			
			PG$^2$~\cite{ma2017pose}  & 28.2403  & 3.5 & 24.2093 & 2.6319  & 26.5138 & 2.8 & 31.7333 & 3.0933 \\ 
			SAMG~\cite{yan2017skeleton} & 28.0185 & 2.6 & 31.2841 & 2.7453 & 26.9545 & 2.3 & 38.1758 & 3.1006 \\ 
			DPIG~\cite{ma2017disentangled}  & 30.6487  & 7.1 &6.7661 & 2.6184 & 26.9451 & 6.9 & 26.2713 & 3.0846 \\  
			PoseGAN~\cite{siarohin2018deformable} & 29.5471 & 9.3 & 9.6725 & 2.5846 & 27.3014 & 8.6 & 24.6712 & 3.0467 \\ 
			GestureGAN~\cite{tang2018gesturegan}  &32.6091 & 26.1  & 7.5860 &2.5223  &27.9749 & 22.6 &18.4595 &2.9836 \\ 
			%			Ours &\textbf{32.6574}& 2.3783 & \textbf{29.3}& \textbf{6.7493} & \textbf{1.7401} & \textbf{31.5420} & 2.2159 & \textbf{27.6}& \textbf{12.4465} & \textbf{2.2104} \\ \hline
			%			Ours & -&\textbf{32.6574} & \textbf{29.3}& \textbf{6.7493} & \textbf{1.7401} & \textbf{31.5420} & \textbf{27.6}& \textbf{12.4465} & \textbf{2.2104} \\ \hline
			%		AsymmetricGAN Heavy &32.5740 &2.3636& & 6.8711 & 1.7519 \\
			%			AsymmetricGAN & Ours &\textbf{32.6686} & \textbf{29.7} & \textbf{6.7132} & \textbf{1.7341} & \textbf{31.5624} & \textbf{28.1} & \textbf{13.1426} & \textbf{2.2311} \\ 	
			AsymmetricGAN &\textbf{32.6686} & \textbf{29.7} & \textbf{6.7132} & \textbf{1.7341} & \textbf{31.5624} & \textbf{28.1} & \textbf{12.4326} & \textbf{2.2011} \\ 	 \bottomrule
	\end{tabular}}
	\label{tab:gesture_comp}
\end{table}

\subsection{Hand Gesture-to-Gesture Translation}
Besides unsupervised image translation tasks, we also conduct experiments on supervised image translation, i.e., hand gesture-to-gesture translation, to validate the effectiveness of the proposed AsymmetricGAN. 	

\noindent\textbf{Datasets and Parameter Settings..}
We follow GestureGAN~\cite{tang2018gesturegan} and employ the NTU Hand Digit~\cite{ren2013robust} and Creative Senz3D~\cite{memo2016head} datasets to evaluate the proposed AsymmetricGAN.
%The number of train/test image pair for NTU Hand Digit and Creative Senz3D are 75,036/9,600 and 135,504/12,800, respectively.
%Each image pair consists of two images of the same person but different gestures.
%All images in our experiments are scaled to $256{\times}256$.
The batch size is set to 4 for both datasets and all the models are trained with 20 epochs. 
Moreover, we follow~\cite{ma2017pose,tang2018gesturegan} and employ OpenPose~\cite{simon2017hand} to extract hand skeletons as training data.
%For both datasets. we do left-right flip and random crops for data augmentation in our experiments.

\noindent\textbf{Evaluation Metric.} 
We follow GestureGAN~\cite{tang2018gesturegan} and employ PSNR, FID~\cite{heusel2017gans}, FRD~\cite{tang2018gesturegan} and AMT user study to evaluate the quality of generated images.
PSNR evaluates generated images and real images from a pixel-level similarity.
Both FID and FRD evaluate generated images and real images from a high-level feature space.
Specifically, FID is a measure of similarity between generated images and real images, and FRD is a measure of distance between each generated image and the corresponding real image.

\begin{figure}[!tbp]  \small
	\centering 
	\includegraphics[width=0.8\linewidth]{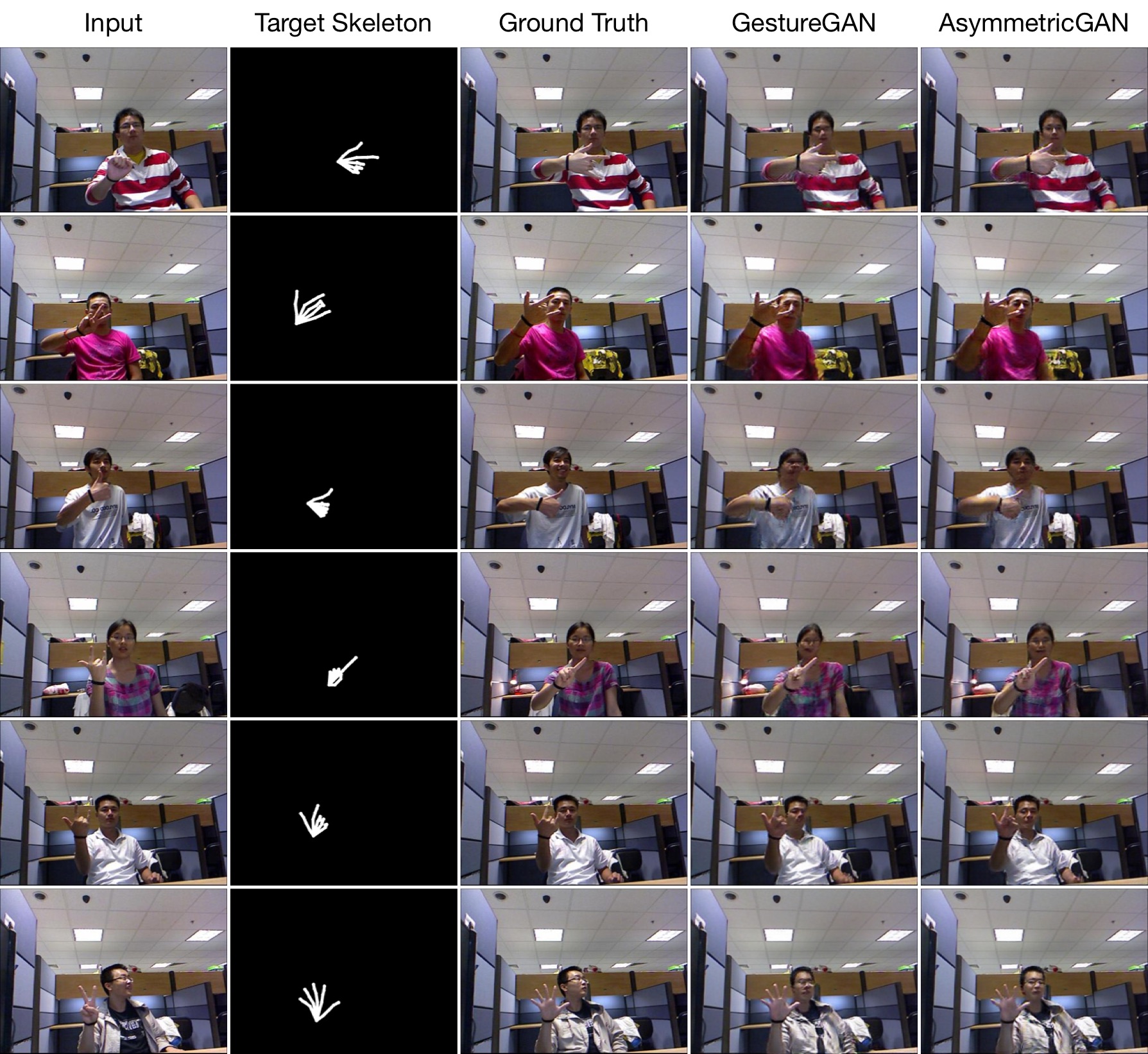}
	\caption{Different methods for hand gesture-to-gesture translation on NTU Hand Digit. 
	\hao{From left to right: Input, Target Skeleton, Ground Truth, GestureGAN~\cite{tang2018gesturegan}, and AsymmetricGAN (Ours).}
	}
	\label{fig:comparison_ntu}
\end{figure}

\noindent\textbf{Ablation Study.}
We conduct ablation studies between SymmetricGAN (i.e., GestureGAN) and AsymmetricGAN on NTU Hand Digit to validate our motivation of the asymmetric network design.
1) SymmetricGAN has two separate generators with the identity structure for both generators $G^t$ and $G^r$, which has 11.388M*2=22.776M parameters totally. 2) AsymmetricGAN also has two separate generators for both generators $G^t$ and $G^r$. However, the filters in first convolutional layer of $G^t$ and $G^r$ are 64 and 4, respectively. It means $G^t$ and $G^r$ has 11.388M and 0.046M parameters, respectively.

Both network architectures are described in Sec.~\ref{sec:for} and comparison results of both generation performance and network parameters are reported in Table~\ref{tab:gesture_abla}.
We see that although the total number of parameters of AsymmetricGAN is much less than SymmetricGAN, AsymmetricGAN still achieves better results than SymmetricGAN on all metrics, which validate our motivation of the asymmetric generator design.

\begin{figure}[!tbp]  \small
	\centering 
	\includegraphics[width=0.8\linewidth]{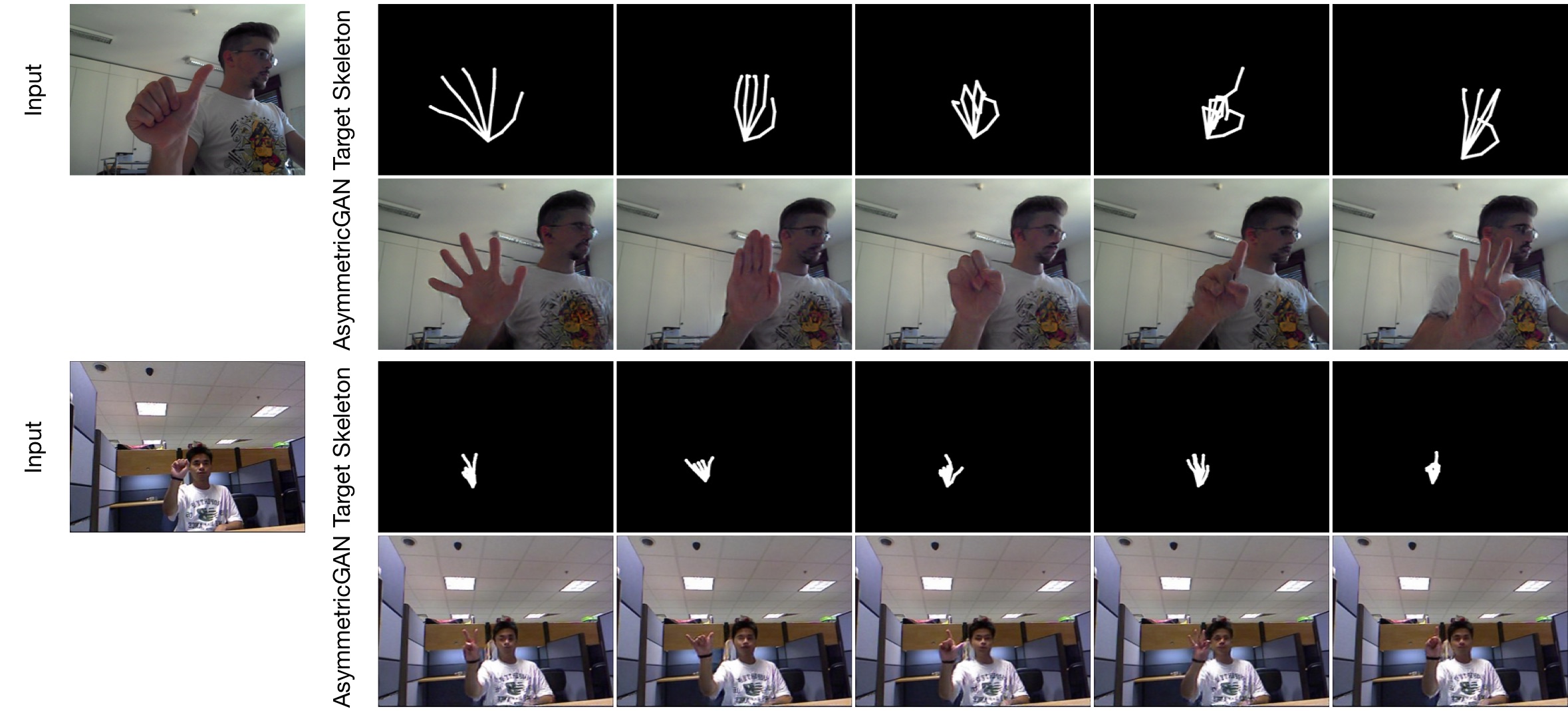}
	\caption{Arbitrary hand gesture translation on NTU Hand Digit (top) and Senz3d (bottom). 
		%		From left to right: Input, Target Skeleton and AsymmetricGAN (ours).
	}
	\label{fig:ab_result}
\end{figure}

\noindent\textbf{State-of-the-Art Comparisons.}
We compare the proposed AsymmetricGAN with the most related five works, i.e., GestureGAN~\cite{tang2018gesturegan}, PG$^2$~\cite{ma2017pose}, DPIG~\cite{ma2017disentangled}, PoseGAN~\cite{siarohin2018deformable} and SAMG~\cite{yan2017skeleton}.
%More specifically, PG$^2$ and DPIG try to produce hand gesture images conditioned on input images and target human keypoints.
%PoseGAN and SAMG use human skeleton information to generate person images.
%GestureGAN is the first work to solve the task of hand gesture-to-gesture translation, which receives the input image and the target hand skeleton to generate the corresponding target hand image.
%Note that SAMG is proposed to generated image sequences, we re-implement this model to produce a single image for a fair comparison.
%All these five methods and the proposed AsymmetricGAN are paired image-to-image translation models.
Quantitative Results of PSNR, FID and FRD are shown in Table~\ref{tab:gesture_comp}.
We see that our performance are significantly much better than existing models on all metrics.
%which equips a parameter-sharing generator for both image translation and image reconstruction.
%For NTU Hand Digit, we can see that the proposed method achieves better results than GestureGAN on PSNR, FID and FRD metrics with 0.0595, 0.8728 and 0.7879 improvements, respectively.
%For Senz3d, we observe that AsymmetricGAN achieves better results than GestureGAN on PSNR, FID and FRD metrics with significantly 3.5875, 6.0269 and 0.7825 improvements, respectively.
Moreover, we visually compare the proposed method with the most related work GestureGAN in Fig.~\ref{fig:comparison_ntu}.  
We see that AsymmetricGAN produces much more photo-realistic results with convincing details compared with GestureGAN, validating our motivation.
We also provide arbitrary hand gesture translation results on both datasets in Fig. \ref{fig:ab_result}, we observe that the proposed method generates different hand gestures according to the target skeletons.

%\begin{table}[!tbp] \small
%	\centering
%	\caption{Comparison with different models for hand gesture recognition.}
%	\resizebox{0.45\linewidth}{!}{
%		\begin{tabular}{lcc} \toprule
%			Model                                             & NTU Hand Digit   & Senz3D  \\ \midrule
%			real/real                                          & 15.000\%          & 34.343\% \\ 
%			PG$^2$~\cite{ma2017pose}                          & 93.667\%          & 98.737\%  \\ 
%			SAMG~\cite{yan2017skeleton}                      & 95.333\%          & 99.495\%  \\ 
%			DPIG~\cite{ma2017disentangled}       & 95.864\%          & 99.054\% \\ 
%			PoseGAN~\cite{siarohin2018deformable}      & 96.128\%          & 99.549\% \\ 
%			GestureGAN~\cite{tang2018gesturegan}  & 96.667\% & 99.747\% \\ 
%			AsymmetricGAN                         & \textbf{97.333\%}                 & \textbf{99.797\%} \\
%	\end{tabular}}
%	\label{tab:data_aug}
%\end{table}

\noindent\textbf{User Study.} Similar to GestureGAN~\cite{tang2018gesturegan}, we also conduct a user study.
The results compared with existing methods are shown in Table~\ref{tab:gesture_comp}.
We note that AsymmetricGAN consistently achieves better results than other baselines on both datasets.

%Moreover, we directly compare our AsymmetricGAN with the most related one work, i.e., GestureGAN~\cite{tang2018gesturegan}, which adopts one parameter-sharing generator for both image translation and image reconstruction.
%The comparison results are also presented in Table~\ref{tab:gesture_comp}.

%We note that the proposed AsymmetricGAN consistently achieves better results compared with other baselines on both datasets.

\noindent\textbf{Data Augmentation.} 
We use the generated images to improve the performance of a hand gesture classifier as in GestureGAN~\cite{tang2018gesturegan}. 
%The intuition is that if the generated images are realistic, the classifiers trained on both the real images and the generated images will be able to boost classification performance. 
Specifically, we employ a pre-trained ResNet50~\cite{he2016deep} as the classifier. 
For both datasets, we make a split of $70\%/30\%$ between training and testing sets. 
Results compared with existing methods are shown in Table~\ref{tab:data_aug}.
The term `real/real' in Table~\ref{tab:data_aug} represents the result without data augmentation. 
We observe that the performance improves significantly after adding the generated images by different methods. 
Moreover, we can see that AsymmetricGAN achieves the best result compared with other methods, meaning the generated images produced by our method are more photo-realistic.

\begin{table}[!tbp] \small
	\centering
	\caption{Comparison with different models for hand gesture recognition.}
	\resizebox{0.8\linewidth}{!}{
		\begin{tabular}{lccccccc} \toprule
			Model                & real/real &  PG$^2$~\cite{ma2017pose}  & SAMG~\cite{yan2017skeleton} & DPIG~\cite{ma2017disentangled} &  PoseGAN~\cite{siarohin2018deformable}  &  GestureGAN~\cite{tang2018gesturegan} & AsymmetricGAN \\ \midrule
			NTU Hand Digit  & 15.000\% & 93.667\% & 95.333\% & 95.864\% & 96.128\% & 96.667\% & \textbf{97.333\%}                  \\
			Senz3D              & 34.343\% & 98.737\% & 99.495\% & 99.054\% & 99.549\%  & 99.747\% & \textbf{99.797\%} \\ \bottomrule
	\end{tabular}}
	\label{tab:data_aug}
\end{table}

%\noindent\textbf{Generalizability.} 
%As shown in Tables \ref{tab:result_amt}, \ref{tab:result_paint}, \ref{tab:rafd} and \ref{tab:gesture_comp}, the proposed method has achieved the new state-of-the-art results on 6 challenging datasets, i.e., Facades, Alps, Collection Style, RaFD, NTU Hand Digit and Senz3D.
%Moreover, as reported in Table \ref{tab:result_amt}, the proposed method has achieved the second best results on the others, i.e., AR Face and Bu3dfe.
%All this evidence proves the generalizability of our model.
%%%%%%%%%%%%%%%%%%%%%
\section{Conclusion}
\label{sec:con}
We present a novel AsymmetricGAN, a robust and efficient model that can perform both supervised and unsupervised image-to-image translations.
The proposed asymmetric dual generators, allowing for different network architectures and different-level parameter sharing strategy, are designed for the image translation and image reconstruction tasks.
Moreover, we explore jointly using different objective functions to optimize AsymmetricGAN, and thus generating images with better fidelity and high quality. 
Extensive experimental results on different scenarios demonstrate that AsymmetricGAN achieves more photo-realistic results and more modeling capacity than existing methods for both unsupervised and supervised image translation tasks.
Finally, the proposed model and training skills can be easily applied to other GAN frameworks.
In the future, we will focus on the face aging task \cite{suo2009compositional}, which aims to generate facial image with different ages in a continuum.

%\begin{acks}
%This work is partially supported by the Italy-China collaboration project TALENT.
%\end{acks}

\bibliographystyle{ACM-Reference-Format}
\bibliography{sample-base}

%\appendix
%Lorem ipsum dolor sit amet, consectetur adipiscing elit. Morbi
%malesuada, quam in pulvinar varius, metus nunc fermentum urna, id
%sollicitudin purus odio sit amet enim. Aliquam ullamcorper eu ipsum
%vel mollis. Curabitur quis dictum nisl. Phasellus vel semper risus, et
%lacinia dolor. Integer ultricies commodo sem nec semper.
%Etiam commodo feugiat nisl pulvinar pellentesque. Etiam auctor sodales
%ligula, non varius nibh pulvinar semper. Suspendisse nec lectus non
%ipsum convallis congue hendrerit vitae sapien. Donec at laoreet
%eros. Vivamus non purus placerat, scelerisque diam eu, cursus
%ante. Etiam aliquam tortor auctor efficitur mattis.
%Nam id fermentum dui. Suspendisse sagittis tortor a nulla mollis, in
%pulvinar ex pretium. Sed interdum orci quis metus euismod, et sagittis
%enim maximus. Vestibulum gravida massa ut felis suscipit
%congue. Quisque mattis elit a risus ultrices commodo venenatis eget
%dui. Etiam sagittis eleifend elementum.
%
%Nam interdum magna at lectus dignissim, ac dignissim lorem
%rhoncus. Maecenas eu arcu ac neque placerat aliquam. Nunc pulvinar
%massa et mattis lacinia.

\end{document}